\documentclass[journal]{IEEEtran}
\usepackage{color}
\usepackage{amssymb,amsfonts,amsmath,graphicx,subfigure,algorithm,algorithmicx}
\usepackage[noend]{algpseudocode}

\newtheorem{definition}{Definition}

\begin{document}
\title{KDCTime: Knowledge Distillation with Calibration on InceptionTime for Time-series Classification}

\author{Xueyuan Gong,
        Yain-Whar Si,
        Yongqi Tian,
        Cong Lin,
        Xinyuan Zhang,
        and Xiaoxiang Liu%\IEEEmembership{Member,~IEEE,}% <-this % stops a space
        \thanks{This work was supported in part by the Fundamental Research Funds for the Central Universities under Grant 21621017, in part by the Innovative Youth Program of Guangdong University under Grant 2019KQNCX194. \textit{(Corresponding author: Xiaoxiang Liu)}}% <-this % stops a space
		\thanks{X. Gong, C. Lin, X. Zhang and X. Liu are with School of Intelligent Systems Science and Engineering, Jinan University, Zhuhai, China (e-mail: xygong@jnu.edu.cn; conglin@jnu.edu.cn; Zhangxy@jnu.edu.cn; tlxx@jnu.edu.cn).}% <-this % stops a space
		\thanks{Y.-W. Si is with Department of Computer and Information Science, University of Macau, Macau, China (e-mail: fstasp@um.edu.mo).}%
		\thanks{Y. Tian is with School of Optoelectronics, Beijing Institute of Technology, Beijing, China (e-mail: 3220190462@bit.edu.cn).}
}

\maketitle

\begin{abstract}
Time-series classification approaches based on deep neural networks are easy to be overfitting on UCR datasets, which is caused by the few-shot problem of those datasets. Therefore, in order to alleviate the overfitting phenomenon for further improving the accuracy, we first propose Label Smoothing for InceptionTime (LSTime), which adopts the information of soft labels compared to just hard labels. Next, instead of manually adjusting soft labels by LSTime, Knowledge Distillation for InceptionTime (KDTime) is proposed in order to automatically generate soft labels by the teacher model. At last, in order to rectify the incorrect predicted soft labels from the teacher model, Knowledge Distillation with Calibration for InceptionTime (KDCTime) is proposed, where it contains two optional calibrating strategies, i.e. KDC by Translating (KDCT) and KDC by Reordering (KDCR). The experimental results show that the accuracy of KDCTime is promising, while its inference time is two orders of magnitude faster than ROCKET with an acceptable training time overhead.
\end{abstract}


% \fi
\IEEEpeerreviewmaketitle

\section{Introduction}
\label{sc:intro}
\begin{figure}[!t]
	\centering
	\subfigure[The gap between training and test accuracy on $113$ UCR datasets]{
		\includegraphics[width=0.45\columnwidth]{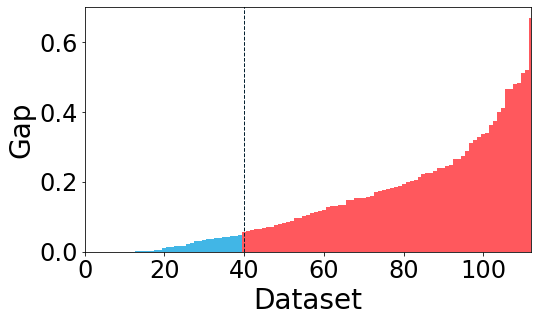}
	}
	\hfil
	\subfigure[The training and test accuracy w.r.t. epochs on $Crop$ dataset]{
		\includegraphics[width=0.45\columnwidth]{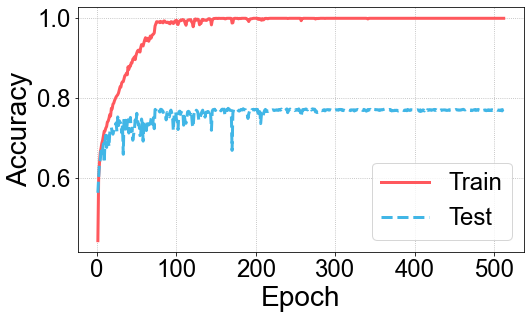}
	}
	\caption{Two examples showing that InceptionTime, including other DNNs, is easy to be overfitting on UCR datasets}
	\label{fg:gap}
\end{figure}

\IEEEPARstart{T}{ime-series} Classification (TSC) is one of the most challenging tasks in data mining \cite{Fawaz:2019}. In recent years, with the remarkable success of deep neural networks (DNNs) in Computer Vision (CV) \cite{Hu:2020, He:2016, Szegedy:2015}, many researchers have tried to employ DNNs in TSC due to the similarity between time-series data (One-dimensional sequence) and image data (Two-dimensional sequence). However, with extensive experiments on various existing DNN-based TSC approaches, we found that DNNs are easy to be overfitting on datasets from the UCR archive \cite{UCRArchive}. To be specific, several experiments are conducted for $3$ typical DNNs, i.e. Fully Convolutional Networks (FCN) \cite{Wang:2017}, Residual Networks (ResNet) \cite{He:2016, Wang:2017}, and InceptionTime \cite{Fawaz:2020}, on the UCR datasets. Selecting InceptionTime as an example, only $40$ datasets have the training and test accuracy gap less than $0.05$, which means the gap on the other $73$ datasets is more than $0.05$, where many of them are more than $0.2$, as shown in Fig. \ref{fg:gap}(a). In addition, Fig. \ref{fg:gap}(b) gives the training and test accuracy with respect to epochs on a specific dataset among UCR datasets, i.e. $Crop$ dataset, where the gap becomes $0.22$ around epoch $100$ and keeps stationary till the end.

After thoroughly analyzing the UCR datasets, we claim that the difference between datasets in CV and those in UCR archive contributes the most to the overfitting phenomenon. In detail, we can view this from the perspective of $N$-shot learning, where $N$ represents the training examples per class. In CV, datasets always contain enough training samples for DNNs, e.g. MNIST \cite{LeCun:1998} and CIFAR10 \cite{CIFAR10} are both $5000$-shot learning datasets with $50000$ training samples in total, and ImageNet \cite{Deng:2009} is averagely a $700$-shot learning dataset with more than 14 million traning samples. However, in UCR archive, $95$ datasets are less than $100$-shot and $120$ datasets with less than $1000$ training samples, while only $4$ ones with more than $2000$ training samples. For example, \textit{Fungi} is a $1$-shot learning dataset, and \textit{DiatomSizeReduction} contains only $16$ training samples. We conclude this as the few-shot problem in TSC. In summary, solving the few-shot problem is the key to improve accuracy. Interestingly, many state-of-the-art approaches adopted methods for alleviating overfitting without pointing out the overfitting problem, such as the Hierarchical Vote system for Collective Of Transformation-based Ensembles (HIVE-COTE) \cite{Lines:2018} with an ensemble of $37$ classifiers, InceptionTime \cite{Fawaz:2020} with an ensemble of $5$ models, RandOm Convolutional KErnel Transform (ROCKET) \cite{Dempster:2020} with $L2$ regularization and Cross Validation, etc. Among the aforementioned state-of-the-art methods, ROCKET has the best accuracy and inference time balance in practice.

%In addition, the DNNs in CV is targeted at only one dataset, yet the DNNs in TSC is targeted at more than $100$ UCR datasets. It is hard to ensure the performance of DNNs with one model and one set of hyperparameters, considering the different data distribution and data characteristic of distinct datasets. We conclude this as the multi-dataset problem in TSC.
	
\begin{figure}[!t]
	\centering
	\subfigure[Soft labels for CV]{
		\includegraphics[width=0.45\columnwidth]{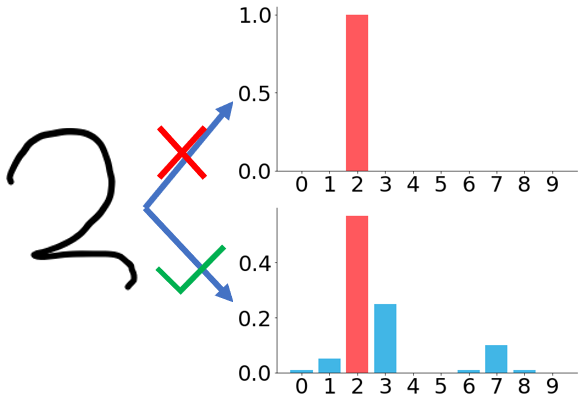}
	}
	\hfil
	\subfigure[Soft labels for TSC]{
		\includegraphics[width=0.45\columnwidth]{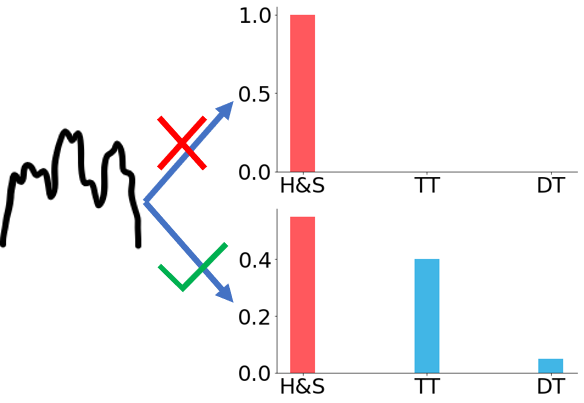}
	}
	\caption{Two examples demonstrating the benefits of soft labels}
	\label{fg:ls}
\end{figure}

In this paper, instead of employing the ordinary approaches for alleviating overfitting, e.g. $L1$ and $L2$ regularization, Batch Normalization (BN) \cite{Ioffe:2015}, Dropout \cite{Srivastava:2014}, early stopping, etc, we first proposed Label Smoothing based on InceptionTime \cite{Szegedy:2016} (LSTime) for improving the generalization ability of InceptionTime, as soft labels are closer to real-life compared to hard labels. For instance, as shown in Fig. \ref{fg:ls}(a), the true label of that handwritten number is $2$. Yet, it also looks like $3$ intuitively. Thus, giving a hard label $2$ to that number may cause information losses. Also, in stocks, there are many meaningful chart patterns, among which Head\&Shoulders (H\&S), Triple Top (TT), and Double Top (DT) \cite{Wan:2017} are similar. In Fig. \ref{fg:ls}(b), the H\&S chart pattern is close to TT. Therefore, we wish to keep its information of TT. As a consequence, soft labels maintain more information compared to hard labels. However, we would like to get the soft labels automatically rather than to set that manually. Thus, secondly, Knowledge Distillation based on InceptionTime \cite{Hinton:2015} (KDTime) is leveraged to generate the soft labels by a teacher model, which is a pre-trained network with a deep and complex architecture. The predicted labels from the teacher model represent the knowledge learned by it. After that, the knowledge (Predicted soft labels) can be distilled to help the training of a student model, which owns a relatively smaller and simpler architecture. As a consequence, Knowledge Distillation can also reduce the inference time because of the student model. At last, we claim that the teacher is not $100\%$ correct, which means it may produce wrong soft labels and misguide the student. As the ground-truth labels have already been obtained, we propose to simply calibrate the wrong predicted soft labels, in order to maximize the accuracy of InceptionTime, called Knowledge Distillation with Calibration based on InceptionTime (KDCTime). In addition, KDCTime includes two optional calibrating strategies, i.e. KDC by Translating (KDCT) and KDC by Reordering (KDCR).

The InceptionTime model employed in LSTime, KDTime, and KDCTime is a single model instead of $5$ ones, since the training and inference time are essential, which indicate the feasibility of that model. Thus, compared to an ensemble of $5$ models with $6$ Inception modules each in its original version, the InceptionTime model in this paper is only one with $3$ Inception modules. In summary, the main contributions in this paper can be concluded as follows:

\begin{itemize}
	\item Discovered the TSC approaches based on DNNs are normally overfitting on the UCR datasets, which is caused by the few-shot problem of those datasets. Thus, the best direction to improve the performance of DNNs is alleviating overfitting.
	\item Combined Label Smoothing and Knowledge Distillation with InceptionTime respectively, denoted LSTime and KDTime. LSTime trains the InceptionTime model with manually controlled soft labels, while KDTime is able to generate soft labels automatically by the teacher model.
	\item Proposed KDCTime for calibrating incorrect soft labels predicted by the teacher model, where it contains two optional strategies, i.e. KDCT and KDCR. As a consequence, KDCTime further improves the accuracy of KDTime.
	\item We have tested the accuracy, training time and test time of ROCKET, InceptionTime, LSTime, KDTime, KDCTime. The results show that compared to ROCKET, KDCTime simultaneously improves the accuracy and reduces the inference time of it with an acceptable training time overhead. As a conclusion, the performance of KDCTime is promising.
\end{itemize}

The remainder of this paper is organized as follows: The related work is reviewed in Section \ref{sc:rw}. Next, LSTime, KDTime, and KDCTime are introduced in Section \ref{sc:pa}. The experimental results are discussed in Section \ref{sc:exp}. Finally, Section \ref{sc:conc} concludes the paper.

\section{Related Work}
\label{sc:rw}
TSC, as a traditional time-series mining research direction, has been considered as one of the most challenging problems \cite{Fawaz:2019}. Traditionally, $1$ Nearest Neighbor (1-NN) Classifiers based on Dynamic Time Warping (DTW) distance has been shown to be a very promising approach \cite{Bagnall:2017}. Yet, the time complexity of DTW is unacceptable compared to Euclidean Distance (ED), i.e. $\mathcal{O}(n^2)$ compared to $\mathcal{O}(n)$. Thus, many researchers have tried to accelerate the execution time of DTW. Rakthanmanon et al. \cite{Rakthanmanon:2012} proposed UCR-DTW by leveraging lower bounding and early abandoning. Sakurai et al. \cite{Sakurai:2007} proposed SPRING under the DTW distance, which is able to monitor time-series streams in real-time. Gong et al. \cite{Gong:2018} proposed Forward-Propagation NSPRING (FPNS) for further accelerate the speed of SPRING. Nevertheless, those researches are all concentrating on time complexity. The upper bound of their accuracy is the accuracy of DTW distance.

In order to breakthrough the bottleneck of accuracy, researchers found that ensembling of several classifiers could significantly improve the accuracy of TSC. Thus, Baydogan et al. \cite{Baydogan:2013} selected an ensemble of decision trees. Kate \cite{Kate:2016} employed an ensemble of several $1$-NN classifiers with different distance measures, including DTW distance. Those methods motivated the development of an ensemble of $35$ classifiers, named Collective Of Transformation-based Ensembles (COTE) \cite{Bagnall:2016}, which ensembles those classifiers over different time-series representations instead of same ones. After that, Lines et al. \cite{Lines:2018} extended COTE by leveraging a new hierarchical structure with probabilistic voting, called HIVE-COTE, which is currently considered the state-of-the-art approach in accuracy. Nevertheless, in order to achieve such a high accuracy, those methods sacrifice the training and inference time, most of which are impractical when datasets are large.

With the rapid development of deep learning, DNNs are widely applied in CV and are also increasingly employed in TSC \cite{Wang:2017}. Cui et al. \cite{Cui:2016} proposed Multi-Scale Convolutional Neural Networks (MCNN), which transforms the time-series into several feature vectors and feeds those vectors into a CNN model. Wang et al. \cite{Wang:2017} implemented several DNN models originated from CV, i.e. MultiLayer Perceptrons (MLP), Fully Convolutional Networks (FCN), and Residual Networks (ResNet), in order to test their performance in TSC, which provides a strong baseline of DNN-based approaches. Based on a more recent DNN structure, i.e. Inception module \cite{Szegedy:2015}, Karimi-Bidhendi et al. \cite{Karimi-Bidhendi:2018} proposed an approach to transform time-series into feature maps using Gramian Angular Difference Field (GADF), and finally feed those maps to an InceptionNet which is pre-trained for image recognition. By extending a more recent version of IncpetionNet, i.e. Inception-V4 \cite{Szegedy:2017}, Fawaz et al. \cite{Fawaz:2020} proposed InceptionTime, which ensembles $5$ Inception-based models to get a promising accuracy. Multi-scale Attention Convolutional Neural Network (MACNN) \cite{Chen:2021} adopted attention mechanism to further improve the accuracy of MCNN. Instead of utilizing convolutions as parts of the model, RandOm Convolutional KErnel Transform (ROCKET) \cite{Dempster:2020} employed random convolutional kernels as feature extractors converting time-series into feature vectors, which were later fed to a ridge regressor. To the best of authors' knowledge, ROCKET owns the best accuracy and inference time balance, while other DNN-based methods always requires a longer training and inference time. Actually, DNN-based methods always suffer from long training time, e.g. MACNN requires $3$ days of running time for training $85$ datasets from the UCR archive. That builds a huge barrier for researchers to reimplement the approach.

We found that InceptionTime is a quite competitive approach. Using only $1$ InceptionTime model instead of ensembling $5$ ones, it owns a slower yet acceptable training time and a $2$ magnitude less inference time compared to ROCKET. Therefore, when InceptionTime only includes $1$ model instead of ensembling several models, we would like to ensure the accuracy of it by the information obtained from soft labels. The idea of utilizing soft labels was proposed by Szegedy et al. \cite{Szegedy:2016}, that controls the smooth level of soft labels by manually setting a parameter $\varepsilon$. Yet, a better way to determine the smooth level of soft labels is generating soft labels automatically by a teacher model, and training a student model with those labels, the idea of which comes from Knowledge Distillation (KD) \cite{Hinton:2015}. Since then, many extensions of KD were proposed. Some researches \cite{Romero:2014, Yim:2017} concentrated on letting the student model learn the feature maps, instead of soft labels, of the teacher model. In addition, Svitov et al. \cite{Svitov:2020} leveraged the predicted labels by the teacher model as class centers, instead of soft labels, guiding the training of the student model. Oki et al. \cite{Oki:2020} integrated KD into triplet loss and utilized the predicted labels as anchor points for guiding the training of the student model. In this paper, instead of employing other types of KD methods, we still concentrated on label-based KD approaches in order to save more execution time. At last, inspired by \cite{Cho:2019, Mirzadeh:2020}, the gap between the student model and the teacher model should be small, or the student model would hard to mimic the teacher model. Therefore, in this paper, a $3$-layer student InceptionTime model is selected as the student model, and a $6$-layer teacher model is employed as the teacher model. 

\section{Proposed Approaches}
\label{sc:pa}
In this section, instead of concentrating on the model, all the proposed approaches are essentially centered with loss functions and labels. First, notations and definitions are given in Section \ref{ssc:nad}. After that, InceptionTime is briefly introduced in Section \ref{ssc:it}. Then, Label Smoothing for InceptionTime (LSTime) is demonstrated in Section \ref{ssc:lsit}. Next, Knowledge Distillation for InceptionTime (KDTime) is depicted in Section \ref{ssc:kdit}. At last, Knowledge Distillation with Calibration for InceptionTime (KDCTime) is illustrated in in Section \ref{ssc:kdcit}, where it contains $2$ strategies, i.e. Calibration by Translating (CT) and Calibration by Reordering (CR).

\subsection{Notations and Definitions}
\label{ssc:nad}
\begin{definition}
	A time-series $\mathbf{x}\in\mathbb{R}^{N}$ is defined as a vector, where $x_{i}$ represents the $i$-th value of $\mathbf{x}$. The corresponding class of $\mathbf{x}$ is a scalar $c\in\{1,2,\dots,C\}$, with $C$ classes in total.
\end{definition}

\begin{definition}
	A class label of $\mathbf{x}$ is defined as a vector $\mathbf{y}=\{y_{1}, y_{2},\dots, y_{C}\}$, where $y_{i}\in[0,1]$ represents the probability of $\mathbf{x}$ belonging to class $c$. In addition, Eq. \eqref{eq:sumy} always holds for any $\mathbf{y}$.
\end{definition}

\begin{equation}
	\label{eq:sumy}
	\sum_{i=1}^{C}y_{i}=1
\end{equation}

\begin{definition}
	The true label of $\mathbf{x}$ is defined as an one-hot vector $\mathbf{y}^{h}$, where all $y^{h}_{i}=0$ except $y^{h}_{c}=1$, called the hard label. The equation of $\mathbf{y}^{h}$ is given in Eq. \eqref{eq:y}.
\end{definition}

\begin{equation}
\label{eq:y}
	y^{h}_{i}=
	\begin{cases}
		1,\hfill \text{if}~i=c\\
		0,\hfill \text{if}~i\neq c
	\end{cases}
\end{equation}

\begin{definition}
	A dataset $\mathbf{D}$ is a pair of sets, including a set of time-series $\mathbf{X}=\{\mathbf{x}_{1}, \mathbf{x}_{2},\dots, \mathbf{x}_{M}\}$ and a set of true labels $\mathbf{Y}^{h}=\{\mathbf{y}^{h}_{1}, \mathbf{y}^{h}_{2},\dots, \mathbf{y}^{h}_{M}\}$ respectively, where each time-series $\mathbf{x}_{i}$ corresponds to a true label $\mathbf{y}^{h}_{i}$.
\end{definition}

\begin{definition}
	An InceptionTime model is treated as a function $\mathcal{F}\in\mathbb{F}$ mapping an input $\mathbf{x}$ into an output $\mathbf{z}\in\mathbb{R}^{C}$, where $\mathbb{F}$ represents the hypothesis space, i.e. the space containing all possibilities of $\mathcal{F}$.
\end{definition}

\begin{definition}
	The predicted label $\mathbf{\hat{y}}$ is produced by normalizing $\mathbf{z}$ with the Softmax function (Eq. \eqref{eq:sm}). Thus, Eq. \eqref{eq:sumy} always holds.
\end{definition}

\begin{equation}
\label{eq:sm}
	\sigma(z_{i})=\hat{y}_{i}=\frac{e^{z_{i}}}{\sum_{j=1}^{C}e^{z_{j}}}
\end{equation}

\begin{definition}
	A loss function $\mathcal{L}$ is a function measuring the difference between the predicted label $\mathbf{\hat{y}}$ and the true label $\mathbf{y}^{h}$, in order to determine the performance of $\mathcal{F}$.
\end{definition}

\begin{definition}
	The problem in the paper is defined as follows: Given a dataset $\mathbf{D}$, find an $\mathcal{F}^{\star}$ minimizing the predefined $\mathcal{L}$. Formally, it is demonstrated in Eq. \eqref{eq:prob}.
\end{definition}

\begin{equation}
\label{eq:prob}
	\mathcal{F}^{\star}=\mathop{\arg\min}_{\mathcal{F}\in\mathbb{F}}\sum_{i=1}^{M}\mathcal{L}(\mathbf{y}^{h}_{i},\sigma(\mathcal{F}(\mathbf{x}_{i})))
\end{equation}

To this end, important notations are briefly summarized in Table \ref{tb:NaD}.

%To this end, we claim that the main contributions in the paper are essentially concentrated in loss functions.
	
\begin{table}[!t]
	\centering
	\caption{Notations and definitions}
	\label{tb:NaD}
	\begin{tabular}{cc}
		\hline
		\textbf{Notations} & \textbf{Definitions} \\
		\hline
		$\mathbf{x}$ & A time-series \\
		$\mathbf{y}^{h}$ & The true label w.r.t. $\mathbf{x}$ \\
		$\mathbf{D}$ & A dataset \\
		$\mathbf{X}$ & A set of time-series in $\mathbf{D}$ \\
		$\mathbf{Y}$ & A set of labels in $\mathbf{D}$ \\
		$\mathcal{F}$ & An InceptionTime model \\
		$\mathbf{z}$ & The output of $\mathcal{F}(\mathbf{x})$ \\
		$\sigma$ & The Softmax function \\
		$\mathbf{\hat{y}}$ & The predicted label of $\sigma(\mathbf{z})$ \\
		$\mathcal{L}$ & A loss function \\
		\hline
	\end{tabular}
\end{table}

\subsection{InceptionTime}
\label{ssc:it}
The ordinary Softmax Cross Entropy Loss is adopted in InceptionTime \cite{Fawaz:2020}. We first implemented the one model version of InceptionTime with $3$ Inception modules, denoted $\mathcal{F}_{S}$. Thus, the loss function of $\mathbf{\hat{y}}=\sigma(\mathcal{F}_{S}(\mathbf{x}))$ is given in Eq. \eqref{eq:sce}.

\begin{equation}
\label{eq:sce}
	\mathcal{L}_{CE}(\mathbf{y}^{h},\mathbf{\hat{y}})=-\sum_{i=1}^{C}y^{h}_{i}\log\hat{y}_{i}
\end{equation}
where it is easy to know the final loss $\mathcal{L}_{CE}$ is only related to $y^{h}_{c}$, as all the other $y^{h}_{i}$ are $0$ (Eq. \eqref{eq:y}). In other words, only the result of $\log\hat{y}_{c}$ is survived in the summation. Therefore, for simplicity, Eq. \eqref{eq:sce} can also be written as Eq. \eqref{eq:sces}.
	
\begin{equation}
\label{eq:sces}
	\mathcal{L}_{CE}(\mathbf{y}^{h},\mathbf{\hat{y}})=-\log\hat{y}_{c}
\end{equation}
	
Note that Eq. \eqref{eq:sces} is the reason that one-hot class label $\mathbf{y}$ is called hard label, as it explicitly selects only the probability of $\mathbf{x}$ belonging class $c$, yet ignoring all other probabilities in the loss function. However, in a more realistic scenario, we believe such deterministic case is rare. Hence, as also introduced in Section \ref{sc:intro}, a soft version of $\mathbf{y}^{h}$ is more feasible in this case.

\subsection{Label Smoothing for InceptionTime}
\label{ssc:lsit}
Second, following by the assumption in Section \ref{ssc:it}, we implemented Softmax Cross Entropy with Label Smoothing (LS) \cite{Szegedy:2016}, with $\mathcal{F}_{S}$ as the model. Therefore, the equation of label smoothed $\mathbf{y}^{h}$ is given in Eq. \eqref{eq:yls}, denoted $\mathbf{y}^{l}$.

\begin{equation}
\label{eq:yls}
	y^{l}_{i}=
	\begin{cases}
		(1-\varepsilon)+\varepsilon/C,\hfill \text{if}~i=c\\
		\varepsilon/C,\hfill \text{if}~i\neq c
	\end{cases}
\end{equation}
where $\varepsilon$ is the smoothing coefficient set by users, representing how much the label is smoothed. Note after LS, $y^{l}_{i}$ still satisfies Eq. \eqref{eq:sumy}. Alternatively, Eq. \eqref{eq:yls} can also be written as Eq. \eqref{eq:ylsa}.
	
\begin{equation}
\label{eq:ylsa}
	y^{l}_{i}=(1-\varepsilon)y^{h}_{i}+\frac{\varepsilon}{C}
\end{equation}

As a consequence, the Softmax Cross Entropy loss with LS is given in Eq. \eqref{eq:scels}.
	
\begin{equation}
\label{eq:scels}
	\begin{aligned}
		\mathcal{L}_{LS}(\mathbf{y}^{l},\mathbf{\hat{y}})
		&=-\sum_{i=1}^{C}y^{l}_{i}\log\hat{y}_{i}\\
		&=-\sum_{i=1}^{C}[(1-\varepsilon)y^{h}_{i}+\frac{\varepsilon}{C}]\log\hat{y}_{i}\\
		&=-\sum_{i=1}^{C}(1-\varepsilon)y^{h}_{i}\log\hat{y}_{i}-\sum_{i=1}^{C}\frac{\varepsilon}{C}\log\hat{y}_{i}\\
		&=(1-\varepsilon)\mathcal{L}_{CE}(\mathbf{y}^{h},\mathbf{\hat{y}})+\varepsilon(-\frac{1}{C}\sum_{i=1}^{C}\log\hat{y}_{i})
	\end{aligned}
\end{equation}
where the left part $(1-\varepsilon)\mathcal{L}(\mathbf{y}^{h},\mathbf{\hat{y}})$ represents the loss from hard labels, while the right part $\varepsilon(-\frac{1}{C}\sum_{i=1}^{C}\log\hat{y}_{i})$ means the loss from soft labels. The smoothing coefficient $\varepsilon$ controls the weights of losses from hard labels and soft labels.
	
Yet, manually controlling the smoothed level of labels by $\varepsilon$ is not the best solution, since, except $y^{l}_{c}$, every smoothed label has the same value. Similar to hard labels, this kind of manually controlled soft labels is not practical in the real world. Therefore, generating flexible soft labels by Knowledge Distillation is then proposed.

\subsection{Knowledge Distillation for InceptionTime}
\label{ssc:kdit}
Third, we implemented Knowledge Distillation (KD) to help us generate soft labels in replacement of manually controlling them. Instead of manually setting up the soft labels, Hinton et al. \cite{Hinton:2015} proposed KD to generate soft labels by a teacher model, which owns a cumbersome architecture with a large number of parameters. Intuitively, the teacher model has more potential to capture the knowledge from training data because of its scale. Next, the predicted labels from the teacher model can be regarded as the knowledge learned by it, denoted $\mathbf{y}^{t}$. Thus, $\mathbf{y}^{t}$ is an automatic soft label compared to the manual soft label $\mathbf{y}^{l}$. Note the one model version of InceptionTime with $6$ Inception modules is incorporated as the teacher model, denoted $\mathcal{F}_{T}$.
	
In addition, instead of directly using Softmax Cross Entropy loss, Softmax with a Temperature $\tau$ and Kullback-Leibler (KL) divergence loss are adopted. The equation of Softmax with $\tau$ is given in Eq. \eqref{eq:smt}.
	
\begin{equation}
\label{eq:smt}
	y^{\tau}_{i}=\frac{e^{z_{i}/\tau}}{\sum_{j=1}^{C}e^{z_{j}/\tau}}
\end{equation}
where the Temperature $\tau$ is a parameter for fine-tuning the smoothed level of predicted labels $\mathbf{y}^{t}$ from the teacher model and $\mathbf{\hat{y}}$ from the student model, denoted $\mathbf{y}^{t_{\tau}}$ and $\mathbf{\hat{y}}^{\tau}$. Note $\tau=1$ means the labels keep unchanged, while $\tau<1$ or $\tau>1$ represents the labels are steeper or smoother respectively. For an extreme example, if $\tau\to\infty$, we will have $\forall i, y^{\tau}_{i}=1/C$. To this end, the loss of $\mathbf{y}^{t_{\tau}}$ and $\mathbf{\hat{y}}^{\tau}$ can be measured by KL divergence, the equation of which is given in Eq. \eqref{eq:kld}.
	
\begin{equation}
\label{eq:kld}
	\mathcal{L}_{KL}(\mathbf{y}^{t_{\tau}},\mathbf{\hat{y}}^{\tau})=\sum_{i=1}^{C}y_{i}^{t_{\tau}}\log\frac{y_{i}^{t_{\tau}}}{\hat{y}_{i}^{\tau}}
\end{equation}
	
After that, $\mathcal{L}_{KL}(\mathbf{y}^{t_{\tau}},\mathbf{\hat{y}}^{\tau})$ representing the loss of soft labels and $\mathcal{L}_{CE}(\mathbf{y}^{h},\mathbf{\hat{y}})$ representing the loss of hard labels are incorporated into a whole for training the student model, which has a relatively small architecture with less parameters. This procedure is regarded as distilling the knowledge from a teacher model into a student model, called KD, in order to preserve the accuracy of the teacher model while reduce its time and space complexity. Note $\mathcal{F}_{S}$ is selected as the student model, which is the one model version of InceptionTime with $3$ Inception modules. The equation of KD loss is given in Eq. \eqref{eq:kdsce}.
	
\begin{equation}
\label{eq:kdsce}
	\begin{aligned}
		&\mathcal{L}_{KD}(\mathbf{y}^{h},\mathbf{\hat{y}},\mathbf{y}^{t_{\tau}},\mathbf{\hat{y}}^{\tau})\\
		&=(1-\varepsilon)\mathcal{L}_{CE}(\mathbf{y}^{h},\mathbf{\hat{y}})+\varepsilon\tau^2\mathcal{L}_{KL}(\mathbf{y}^{t_{\tau}},\mathbf{\hat{y}}^{\tau})
	\end{aligned}
\end{equation}
where $\varepsilon$ controls the weight of $\mathcal{L}_{CE}(\mathbf{y}^{h},\mathbf{\hat{y}})$ (Eq. \eqref{eq:sce}) and $\mathcal{L}_{KL}(\mathbf{y}^{t_{\tau}},\mathbf{\hat{y}}^{\tau})$ (Eq. \eqref{eq:kld}). Note $\tau^{2}$ is necessary because the scale of $\mathcal{L}_{KL}(\mathbf{y}^{t_{\tau}},\mathbf{\hat{y}}^{\tau})$ becomes smaller after fine-tuning by $\tau$. Thus, multiplying a $\tau^{2}$ helps it to be the same scale of $\mathcal{L}_{CE}(\mathbf{y}^{h},\mathbf{\hat{y}})$, so that the total loss has no preference to $\mathcal{L}_{CE}(\mathbf{y}^{h},\mathbf{\hat{y}})$.
	
Nevertheless, similar to teachers in real-life, the teacher model is not ensured to be $100\%$ correct. Sometimes it may misguide the student model to wrong answers. To be specific, the incorrect soft labels generated by the teacher model will also result in the wrong labels predicted by the student model. In order to alleviate the affection of incorrect labels, KD adopts $\mathcal{L}_{CE}(\mathbf{y}^{h},\mathbf{\hat{y}})$ and $\tau$. Nonetheless, that brings additional hyperparameters into the model. Therefore, we would like to propose a better method to alleviate the affection of incorrect labels while not bringing additional hyperparameters.

\subsection{Knowledge Distillation with Calibration for InceptionTime}
\label{ssc:kdcit}
\begin{figure}[!t]
	\centering
	\subfigure[The $2$-D label space when $C=2$]{
		\includegraphics[width=0.4\columnwidth]{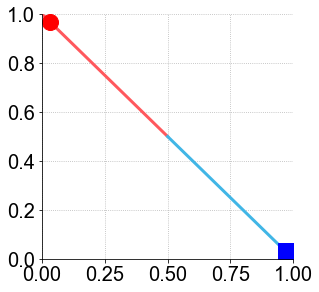}
	}
	\hfil
	\subfigure[The $3$-D label space when $C=3$]{
		\includegraphics[width=0.4\columnwidth]{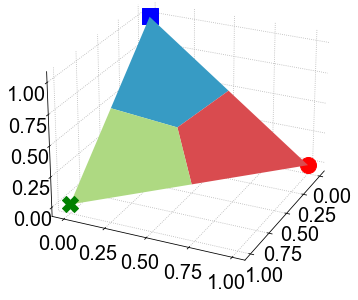}
	}
	\caption{Two examples showing the label spaces when $C=2$ and $C=3$}
	\label{fg:spc}
\end{figure}

At last, we propose Knowledge Distillation with Calibration (KDC) to calibrate the incorrect soft labels generated by the teacher model before distillation. Note the teacher model and student model are $\mathcal{F}_{T}$ and $\mathcal{F}_{S}$ respectively (Section \ref{ssc:kdit}). In this way, it is not necessary to employ $\mathcal{L}_{CE}(\mathbf{y},\mathbf{\hat{y}})$ and $\tau$ in KDC. In order to calibrate the incorrect soft labels, all labels $\mathbf{y}$ are regarded as vectors geometrically, including the hard label $\mathbf{y}^{h}$ and the soft label $\mathbf{y}^{t}$ generated by the teacher model. From this point of view, according to Eq. \eqref{eq:sumy}, the feasible solution space of $\mathbf{y}$ is a triangular hyperplane, named the label space. In other words, all $\mathbf{y}$ are located on a triangular hyperplane. Fig. \ref{fg:spc}(a) gives an example when $C=2$. In this case, the triangular hyperplane is an $1$-D line in $2$-D space. Next, Fig. \ref{fg:spc}(b) shows another example when $C=3$. In this case, the triangular hyperplane is a $2$-D regular triangle in $3$-D space. Last, the triangular hyperplane is a $3$-D regular tetrahedron in $4$-D space when $C=4$. Nonetheless, we failed to plot a $4$-D space in figures. Note that distinct colors represent the areas of distinct classes. Thus, it is potential to calibrate $\mathbf{y}^{t}$ from its original position to the target position $\mathbf{y}^{h}$, if $\mathbf{y}^{t}$ is located in the wrong area. In addition, all $\mathbf{y}^{h}$ are located at the vertices of the triangular hyperplane, as marked in Fig. \ref{fg:spc}(a) and Fig. \ref{fg:spc}(b).

Therefore, our main task is to propose a proper method in order to modify $\mathbf{y}^{t}$ to its correct area while its new position is located between $\mathbf{y}^{t}$ and $\mathbf{y}^{h}$. The calibrated $\mathbf{y}^{t}$ is denoted as $\mathbf{y}^{t_{\zeta}}$. To this end, two approaches for calibration are proposed, which are calibration by translating and calibration by reordering. Note that only incorrect predicted labels will be calibrated. Formally, given a $\mathbf{y}^{t}$ and its corresponding $\mathbf{y}^{h}$, $\mathbf{y}^{t_{\zeta}}$ will be computed only when $\mathop{\arg\max}_{i}\{y^{t}_{i}\}\neq\mathop{\arg\max}_{i}\{y^{h}_{i}\}$. In other words, $\mathop{\arg\max}_{i}\{y^{t}_{i}\}\neq c$.

\subsubsection{Calibration by Translating}
\label{sssc:cbt}
Calibration by Translating (CT) represents geometrically translate $\mathbf{y}^{t}$ from its original position to $\mathbf{y}^{h}$, the equation of which is shown in Eq. \eqref{eq:ctl}.

\begin{equation}
\label{eq:ctl}
	\mathbf{y}^{t_{\zeta}}=\mathbf{y}^{t}+\omega(\mathbf{y}^{h}-\mathbf{y}^{t})
\end{equation}
where $(\mathbf{y}^{h}-\mathbf{y}^{t})$ represents the vector from $\mathbf{y}^{t}$ to $\mathbf{y}^{h}$, while $\omega\in[0,1]$ is a calibration coefficient controlling the distance $\mathbf{y}^{t}$ moves towards $\mathbf{y}^{h}$. It is easy to know that $\mathbf{y}^{t_{\zeta}}=\mathbf{y}^{h}$ when $\omega=1$, and $\mathbf{y}^{t_{\zeta}}=\mathbf{y}^{t}$ when $\omega=0$. In this case, it is simply substitution instead of calibration.

Hence, $\omega$ is the key coefficient defining the degree of calibration. We define the equation of $\omega$ as Eq. \eqref{eq:w}.

\begin{equation}
\label{eq:w}
	\omega=\frac{\delta}{\Vert\mathbf{y}^{h}-\mathbf{y}^{t}\Vert_{2}}
\end{equation}
where $\delta$ is the minimum distance between $\mathbf{y}^{h}$ and $\mathbf{y}^{t}$ when $\mathop{\arg\max}_{i}\{y^{t}_{i}\}\neq\mathop{\arg\max}_{i}\{y^{h}_{i}\}$, and $\Vert\mathbf{y}^{h}-\mathbf{y}^{t}\Vert_{2}$ is the current distance between $\mathbf{y}^{h}$ and $\mathbf{y}^{t}$. It is easy to know that $\Vert\mathbf{y}^{h}-\mathbf{y}^{t}\Vert_{2}\geq\delta$ always holds.

To this end, we calculated $\delta$, and got the magic number $\delta=1/\sqrt{2}$. The procedure of calculation is given in Appendix \ref{ap:pcd}. As a consequence, Eq. \eqref{eq:ctl} can be also rewritten as \eqref{eq:ctlf}.

\begin{equation}
\label{eq:ctlf}
	\mathbf{y}^{t_{\zeta}}=\mathbf{y}^{t}+\frac{1}{\sqrt{2}}\frac{\mathbf{y}^{h}-\mathbf{y}^{t}}{\Vert\mathbf{y}^{h}-\mathbf{y}^{t}\Vert_{2}}
\end{equation}
where we know the unit vector $(\mathbf{y}^{h}-\mathbf{y}^{t})/\Vert\mathbf{y}^{h}-\mathbf{y}^{t}\Vert_{2}$ decides the direction of translating, while $\delta=1/\sqrt{2}$ determines the distance of translating. In this way, it ensures that all $\mathbf{y}^{t_{\zeta}}$ stay in the label space, since it guarantees $\omega\leq1$. In addition, it also guarantees that $\omega\geq0$, which leads to a consequence that $\mathbf{y}^{t}$ will not stay unchanged.

\subsubsection{Calibration by Reordering}
\label{sssc:cbr}
Calibration by Reordering (CR) represents reprioritizing the values of the incorrect predicted label based on a specific strategy. To be concrete, given a $\mathbf{y}^{t}$ and its corresponding $\mathbf{y}^{h}$, some $y^{t}_{i}$ will be resorted if $\mathop{\arg\max}_{i}\{y^{t}_{i}\}\neq\mathop{\arg\max}_{i}\{y^{h}_{i}\}$. Therefore, our main task is to design a reordering strategy.

Given a $\mathbf{y}^{t}$ awaiting for reordering. The reordering strategy is designed as follows: 1) $\mathbf{y}^{t}$ is sorted by descending order. The sorted $\mathbf{y}^{t}$ is denoted $\mathbf{y}^{t_{s}}$. Thus, since the descending order of $y^{t}_{i}$ is random, we have $\mathbf{y}^{t_{s}}=\{y^{t_{s}}_{1}, y^{t_{s}}_{2},\dots, y^{t_{s}}_{C}\}=\{y^{t}_{1st},y^{t}_{2nd},\dots\,y^{t}_{Cth}\}$, where $y^{t}_{1st}$ is the largest $y^{t}_{i}$, $y^{t}_{2nd}$ is the second largest $y^{t}_{i}$, and so on and so forth. Note $\mathbf{y}^{t_{s}}$ is in descending order, which means $y^{t_{s}}_{1}=y^{t}_{1st}=\max_{i}\{y^{t}_{i}\}$. 2) After defining a temporary value $y^{t}_{tmp}=y^{t}_{1st}$, the value of $y^{t}_{(i+1)th}$ will be assigned to $y^{t}_{ith}$, from $y^{t}_{1st}$, $y^{t}_{2nd}$, all the way to $y^{t}_{ith}=y^{t}_{c}$. 3) Assign the value of $y^{t}_{tmp}$ to $y^{t}_{c}$.

The algorithm of the whole procedure is given in Algorithm \ref{ag:cr}. It ensures that $\mathbf{y}^{t_{\zeta}}$ is located in the label space, since $\mathbf{y}^{t_{\zeta}}$ is only the reordered version of $\mathbf{y}^{t}$. In addition, it guarantees that $\mathbf{y}^{t_{\zeta}}$ is geometrically located
in its class area. In other words, $\mathop{\arg\max}_{i}\{y^{t_{\zeta}}_{i}\}=\mathop{\arg\max}_{i}\{y^{h}_{i}\}$. As a consequence, $\mathbf{y}^{t}$ is successfully calibrated by reordering.

\begin{algorithm}[!t]
	\caption{Algorithm of calibration by reordering}
	\label{ag:cr}
	\begin{algorithmic}[1]
		\Require Given a $\mathbf{y}^{t}$ and its corresponding $\mathbf{y}^{h}$
		\Ensure The reordered label $\mathbf{y}^{t_{\zeta}}$
		\State Sort $\mathbf{y}^{t}$ to get $\mathbf{y}^{t_{s}}=\{y^{t}_{1st},y^{t}_{2nd},\dots\,y^{t}_{Cth}\}$
		\State $y^{t}_{tmp}\gets y^{t}_{1st}$
		\For{$ith=1st,2nd,\dots$, until $ith=c$}
		\State $y^{t}_{ith}\gets y^{t}_{(i+1)th}$
		\EndFor
		\State $y^{t}_{c}\gets y^{t}_{tmp}$
	\end{algorithmic}
\end{algorithm}

\subsubsection{Analysis of CT and CR}
\label{sssc:actcr}
We theoretically analyzed the difference between CT and CR in terms of calibration. To be specific, for one $\mathbf{y}^{t}$ that is not located in the correct area in the label space, KDC will be utilized to calibrate $\mathbf{y}^{t}$ as $\mathbf{y}^{t_{\zeta}}$ either by CT or CR, where $\mathbf{y}^{t_{\zeta}}$ is located in the correct area. It can be concluded that there must exist a mapping of any label from the incorrect area to the correct area in the label space. Thus, we wish to know the mapping relationship by calculating the distance between $\mathbf{y}^{h}$ and $\mathbf{y}^{t}$ after KDC.

\begin{figure}[!t]
	\centering
	\subfigure[Calibration by traslating]{
		\includegraphics[width=0.45\columnwidth]{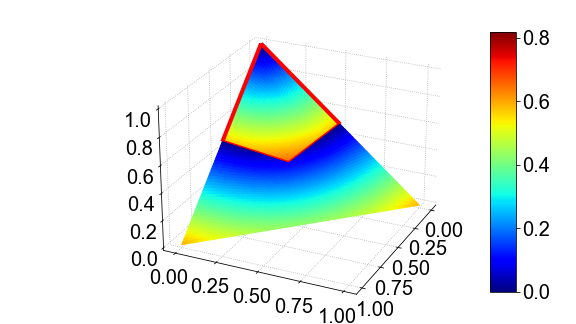}
	}
	\hfil
	\subfigure[Calibration by reordering]{
		\includegraphics[width=0.45\columnwidth]{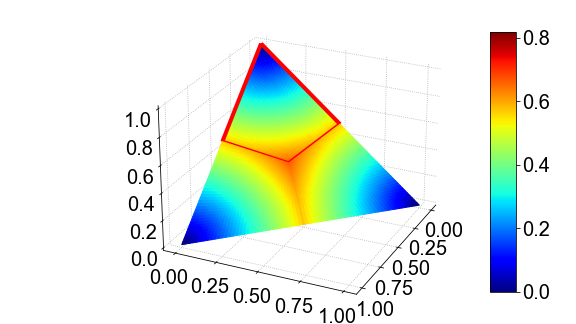}
	}
	\caption{A $3$-D example illustrating the color map of the distance between $\mathbf{y}^{h}$ and $\mathbf{y}^{t}$ after KDC in the label space}
	\label{fg:mctcr}
\end{figure}

Fig. \ref{fg:mctcr} shows an example in $3$-D space, where the color map of the distance between $\mathbf{y}^{h}$ and $\mathbf{y}^{t}$ after KDC in the label space is illustrated. The distance calculating function is defined as $\mathcal{D}(\mathbf{y}^{h},\mathbf{y}^{t})=\Vert\mathbf{y}^{h}-\mathbf{y}^{t}\Vert_{2}$. As $\mathbf{y}^{t}$ in the red area (Correct area) do not require calibration, their distance is calculated by $\mathcal{D}(\mathbf{y}^{h},\mathbf{y}^{t})$ directly. On the contrary, $\mathbf{y}^{t}$ out the red area (Incorrect area) are calibrated as $\mathbf{y}^{t_{\zeta}}$. Therefore, their distance is calculated by $\mathcal{D}(\mathbf{y}^{h},\mathbf{y}^{t_{\zeta}})$. As shown in Fig. \ref{fg:mctcr}(a), the $\mathbf{y}^{t}$ being close to the edge of the red area are mapped close to the vertex, i.e. the hard label $\mathbf{y}^{h}$, while the $\mathbf{y}^{t}$ being far away from the red area are mapped to the edge of the red area. Yet in Fig. \ref{fg:mctcr}(b), the $\mathbf{y}^{t}$ being close to the edge of the red area are mapped also close to the edge, while the $\mathbf{y}^{t}$ being far away from the red area are mapped to close to $\mathbf{y}^{h}$. As a consequence, CT keeps the relative position of incorrect $\mathbf{y}^{t}$ unchanged, which means it preserves the spatial information of $\mathbf{y}^{t}$. In addition, CR keeps the shape of the distribution of incorrect $\mathbf{y}^{t}$ unchanged, which means it preserves the distributional information of $\mathbf{y}^{t}$.

To this end, we are in position to define the loss function of KDC. Unlike the loss function of KD, we employ KL-divergence only without temperature $\tau$ and the cross entropy part, as shown in Eq. \ref{eq:kdc}.
	
\begin{equation}
\label{eq:kdc}
	\mathcal{L}_{KDC}(\mathbf{y}^{t_{\zeta}},\mathbf{\hat{y}})=\mathcal{L}_{KL}(\mathbf{y}^{t_{\zeta}},\mathbf{\hat{y}})=\sum_{i=1}^{C}y_{i}^{t_{\zeta}}\log\frac{y_{i}^{t_{\zeta}}}{\hat{y}_{i}}
\end{equation}
where $\mathbf{y}^{t_{\zeta}}$ is the calibrated label generated by the teacher model, while $\mathbf{\hat{y}}$ is the label generated by the student model. Compared to $\mathcal{L}_{KD}(\mathbf{y}^{h},\mathbf{\hat{y}},\mathbf{y}^{t_{\tau}},\mathbf{\hat{y}}^{\tau})$, $\mathcal{L}_{KDC}(\mathbf{y}^{t_{\zeta}},\mathbf{\hat{y}})$ does not contains any hyperparameter and computes only KL divergence, which reduces much computational time and the complexity of hyperparameter tuning.
	
Thus, the total process of KDC can be concluded as $3$ steps: 1) Train a teacher model with a heavier and more complex architecture by true labels (Hard labels). Generate the predicted labels by the teacher model; 2) Calibrate the incorrect predicted labels; 3) Train the student model with a relatively small and simple architecture by calibrated labels (Soft labels). The algorithm of KDC is given in Algorithm \ref{ag:kdc}.
	
\begin{algorithm}[!t]
\caption{Algorithm of KDCTime}
\label{ag:kdc}
\begin{algorithmic}[1]
	\Require The training data $\mathbf{X}$ and its corresponding labels $\mathbf{Y}$.
	\Ensure The trained student model $\mathcal{F}_{S}^{*}$
	\State Initialize a teacher model $\mathcal{F}_{T}$
	\State Train $\mathcal{F}_{T}$ by $\mathbf{X}$ and $\mathbf{Y}$ to get $\mathcal{F}_{T}^{*}$
	\State Generate $\mathbf{Y}^{t}$ by $\mathcal{F}_{T}^{*}$
	\For{each $\mathbf{y}^{t}_{i}$}
		\If{$\mathbf{y}^{t}_{i}$ and $\mathbf{y}^{h}_{i}$ belong to distinct class}
			\State Calibrate $\mathbf{y}^{t}_{i}$ to get $\mathbf{y}^{t_{\zeta}}_{i}$\Comment{Eq. \ref{eq:ctlf} or Algorithm \ref{ag:cr}}
		\EndIf
	\EndFor
	\State Initialize the student model $\mathcal{F}_{S}$
	\State Train $\mathcal{F}_{S}$ by $\mathbf{X}$ and $\mathbf{Y}^{t_{\zeta}}$ to get $\mathcal{F}_{S}^{*}$
\end{algorithmic}
\end{algorithm}
	
\section{Experiments}
\label{sc:exp}
We conduct the experiments on UCR datasets, in which there are $128$ datasets. Yet, there are $15$ problematic datasets with $NaN$ (Not a Number) values, since data missing or various time-series length. They are listed in Table \ref{tb:pd}. Therefore, the remaining $113$ datasets are selected for experiments.

\begin{table}[!t]
	\centering
	\caption{Problematic Datasets}
	\label{tb:pd}
	\begin{tabular}{ccc}
		\hline
		 & \textbf{Dataset Name} & \\
		\hline
		AllGestureWiimoteX & DodgerLoopDay & GestureMidAirD1 \\
		AllGestureWiimoteY & DodgerLoopGame & GestureMidAirD2 \\
		AllGestureWiimoteZ & DodgerLoopWeekend & GestureMidAirD3 \\
		MelbournePedestrian & PickupGestureWiimoteZ & GesturePebbleZ1 \\
		ShakeGestureWiimoteZ & PLAID & GesturePebbleZ2 \\
		\hline
	\end{tabular}
\end{table}

In the experiments, ROCKET, Softmax cross entropy for InceptionTime (ITime), label smoothing for InceptionTime (LSTime), Knowledge Distillation for InceptionTime (KDTime), and KD with calibration for InceptionTime (KDCTime) are compared, where KDCTime includes two calibrating methods, i.e. KDC by traslating (KDCT) and KDC by reordering (KDCR) respectively. Except ROCKET, all the aforementioned methods can be concluded as ITime-based approaches. At first, in order to find the best hyperparameter for LSTime, KDTime, KDCT, and KDCR, we conducted the experiments of hyperparameter study for those approaches. Note ITime does not have any hyperparameter to be tuned. After that, we tested the accuracy, training time, and test time of the aforementioned approaches.

Similar to \cite{Fawaz:2020}, critical difference diagrams are drawn in the paper for better illustrating the results of different approaches, as the results of $113$ datasets are hard to be depicted clearly. Critical difference diagram is a diagram drawn by the following steps: 1) Execute the Friedman test \cite{Friedman:1940} for rejecting the null hypothesis. 2) Perform the pairwise post-hoc analysis \cite{Benavoli:2016} by a Wilcoxon signed-rank test with Holm's alpha ($5\%$) correction \cite{Garcia:2008}. 3) Visualize the statistical result by \cite{Demsar:2006}, where a thick horizontal line represents that the approaches are not significantly different with respect to results.

Our experiments are conducted on a computer equipped with an Intel Core i$9$-$11900$ CPU at $2.50$GHz, $32$GB memory, and a NVIDIA GeForce RTX $3090$ GPU. The operating system is Windows 10. Additionally, the development environment is Anaconda $4.10.3$ with Python $3.8.8$ and Pytorch $1.9.0$.

\subsection{Hyperparameter Study for ITime-based Approaches}
\label{ssc:hsa}
In this section, we first searched $3$ hyperparameters, i.e. batch size, epoch, and learning rate, on ITime. Thus, those hyperparameters of LSTime, KDTime, KDCT and KDCR can be determined also, since all these methods are based on InceptionTime. After that, $\varepsilon$ in LSTime, $\varepsilon$ and $\tau$ in KDTime, and $\varepsilon$ in KDCT and KDCR were searched separately.

\subsubsection{Batch Size}
\label{sssc:bs}
First, the batch size was set to $64$ without searching. The reason is that batch size is theoretically the larger the better. An extreme case is the full-batch. However, in deep learning tasks, that always leads to a consequence where graphics memory on GPU is infeasible to load the large dataset. Additionally, The batch size and the epoch are depending on each other, i.e. the more batch size represents the more epochs for converging, which means a longer training time. Thus, the batch size is always empirically set to either $16$, $32$, $64$, $128$, or $256$. Hence, $64$ was adopted in the paper.

\subsubsection{Epoch}
\label{sssc:epc}
\begin{figure}[!t]
	\centering
	\includegraphics[width=1.0\columnwidth]{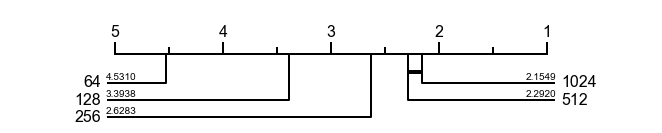}
	\caption{Critical difference diagram for different epochs}
	\label{fg:cdep}
\end{figure}

With a fixed batch size $64$, we compared the accuracy of $5$ different epochs, which were $64$, $128$, $256$, $512$, and $1024$ respectively. The critical difference diagram is given in Fig. \ref{fg:cdep}, where $1024$ epochs owns the best accuracy. Yet, $1024$ epochs is of no critical difference of $512$ epochs. Since the training time of DNN-based approaches is slow, $512$ is selected as the epoch in order to save the training time. In addition, we also employed the early stop strategy with patience equals to $80$ epochs for reducing the training time and alleviating overfitting.

\subsubsection{Learning Rate}
\label{sssc:lr}
\begin{figure}[!t]
	\centering
	\includegraphics[width=1.0\columnwidth]{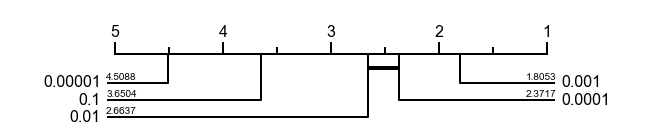}
	\caption{Critical difference diagram for different learning rates}
	\label{fg:cdlr}
\end{figure}

The accuracy of $5$ distinct learning rates were tested in total, which were $0.1$, $0.01$, $0.001$, $0.0001$, and $0.00001$. Moreover, learning rate decay was employed in order to stabilize the training process. We leveraged fixed step decay, also called piecewise constant decay, as the decay strategy, where the step size is set as $35$ and gamma is set as $0.5$. In other words, the learning rate will multiply by $0.5$ for every $35$ epochs. Note there are $256$ epochs, which ensures $7$ times of decay in total. The critical difference diagram is shown in Fig. \ref{fg:cdlr}, where the learning rate being equal to $0.01$ has the highest accuracy. Thus, $0.01$ is employed as the learning rate in the paper.

%\subsubsection{Optimizer}
%\label{sssc:opt}
%\colorbox{yellow}{$XXX$}

\subsubsection{$\varepsilon$ in LSTime}
\label{sssc:lst}
\begin{figure}[!t]
	\centering
	\includegraphics[width=1.0\columnwidth]{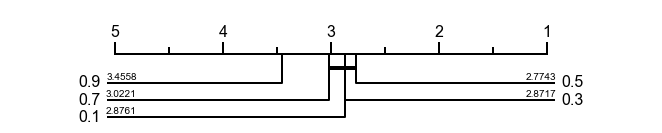}
	\caption{Critical difference diagram for different $\varepsilon$ in LSTime}
	\label{fg:lst}
\end{figure}

The smoothing coefficient $\varepsilon$ (Eq. \eqref{eq:scels}) represents the smoothed level of labels. We tested the accuracy of $5$ different $\varepsilon$ in LSTime, which are $0.1$, $0.3$, $0.5$, $0.7$, and $0.9$. After all, the critical difference diagram is given in Fig. \ref{fg:lst}, where $0.5$ gets the best accuracy. This claims that $\varepsilon$ should not be too small or too big, since a small $\varepsilon$ gives the label little additional information, while a big $\varepsilon$ causes too much information loss from its original class. In addition, the accuracy of big $\varepsilon$ is less than that of small $\varepsilon$, which means information from its original class is important, and it is not a good idea to completely abandon those information.

\subsubsection{$\varepsilon$ and $\tau$ in KDTime}
\label{sssc:kdt}
\begin{figure}[!t]
	\centering
	\includegraphics[width=1.0\columnwidth]{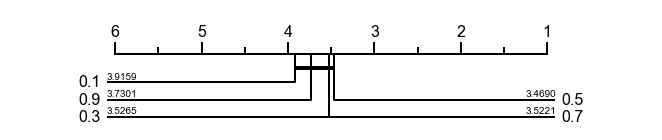}
	\caption{Critical difference diagram for different $\varepsilon$ in KDTime}
	\label{fg:kde}
\end{figure}

\begin{figure}[!t]
	\centering
	\includegraphics[width=1.0\columnwidth]{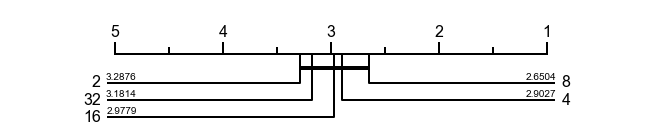}
	\caption{Critical difference diagram for different $\tau$ in KDTime}
	\label{fg:kdt}
\end{figure}

KDTime contains $2$ hyperparameters, which are $\varepsilon$ and $\tau$ (Eq. \eqref{eq:kdsce}), where $\varepsilon$ controls the weight of losses between the hard label and the smooth label respectively, and $\tau$ fine-tunes the smoothed level of smooth labels (Predicted labels by teacher model). Thus, we tested the accuracy of $5$ distinct $\varepsilon$ in KDTime, which are $0.1$, $0.3$, $0.5$, $0.7$, and $0.9$, while we also compared the accuracy of $5$ different $\tau$, which are $2$, $4$, $8$, $16$, and $32$. The critical diagrams of $\varepsilon$ and $\tau$ are shown in Fig. \ref{fg:kde} and Fig. \ref{fg:kdt}. Fig. \ref{fg:kde} shows that $\varepsilon=0.5$ has the best accuracy. The $\varepsilon$ close to $0$ or $1$ will reduce the accuracy. In addition, Fig. \ref{fg:kdt} shows that $\tau=8$ is the best. Similarly, the accuracy of KDTime will decrease if $\tau$ is too big or too small.

To this end, we are able to conclude that all the ITime-based methods select $64$ as batch size, $512$ as epoch, and $0.01$ as learning rate. In addition, $\varepsilon$ in LSTime is set to $0.5$, $\varepsilon$ and $\tau$ in KDTime is set to $0.5$ and $8$ respectively. Finally, Adam is adopted as the optimizing algorithm in order to update the model.

\subsection{Accuracy of Different Approaches}
\label{ssc:aa}
In this section, we compared the accuracy of ROCKET, ITime, LSTime, KDTime, KDCTime by Translating (KDCT) and KDCTime by Reordering (KDCR). The accuracy of all approaches are averaged from $10$ times running. In the mean time, the standard deviation of these $10$ accuracy is also calculated. The result is listed in Table \ref{tb:aa} (Appendix \ref{ap:aa}). The number before the $\pm$ sign represents accuracy, while the number after the $\pm$ sign means standard deviation. Besides, the bold number represents the best accuracy or standard deviation among all approaches.

\begin{figure}[!t]
	\centering
	\includegraphics[width=1.0\columnwidth]{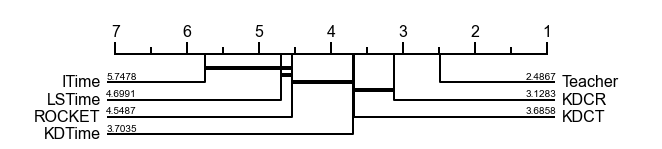}
	\caption{Critical difference diagram illustrating the accuracy of different approaches}
	\label{fg:acc}
\end{figure}

\begin{figure}[!t]
	\centering
	\includegraphics[width=1.0\columnwidth]{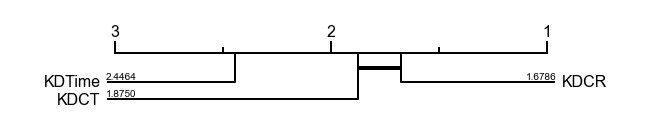}
	\caption{Critical difference diagram illustrating the accuracy of three approaches on datasets with accuracy less than $0.8$ for KDTime}
	\label{fg:acc3}
\end{figure}

As shown in Table \ref{tb:aa}, KDCR gets the highest accuracy on $54$ datasets, which claims that its accuracy is competitive and promising. Yet, its converging process is not as stable as ROCKET, since ROCKET has the lowest standard deviation on the majority of datasets. Besides, the critical difference diagram is also given for better illustrating the result of Table \ref{tb:aa}, which is shown in Fig. \ref{fg:acc}. Note that Fig. \ref{fg:acc} also demonstrates the accuracy of the teacher model used in KDTime, KDCT and KDCR, which is a InceptionTime model with $6$ Inception modules. It is also trained $10$ times and the best one is selected as the teacher. That is the reason why its accuracy is the best one. In addition, KDCR and KDCT are of no significant difference with KDTime. The reason is that the accuracy of $75$ datasets are more than $0.8$ for the teacher model, which claims the majority of datasets meets the bottleneck of improving accuracy. In other words, only a small number of samples can be calibrated by KDCTime. Thus, We claim that the results of those $85$ datasets dominates the other $28$ datasets in the critical difference diagram. By ignoring that part of results, the accuracy of KDCR and KDCT is more promising, as shown in Fig. \ref{fg:acc3}.

In summary, KDCR owns the best accuracy, which is better than KDCT. The reason is that KDCR keeps the information from marginal labels. In Fig. \ref{fg:spc}, marginal labels represent the labels located in the middle area of the label space. In Fig. \ref{fg:mctcr}(b), we know the labels located in the middle area have a long distance to $\mathbf{y}^{h}$, where they do not deterministically belong to any class, meaning that they contain abundant information from other classes. Yet, KDCT calibrates the marginal labels close to $\mathbf{y}^{h}$, which losses those information. In addition, KDCR calibrates the labels close to other classes as close to the correct one, as shown in Fig. \ref{fg:mctcr}(b), which eliminates the misguiding from deterministic incorrect labels. Nevertheless, KDCT keeps the information of those labels from incorrect classes.

\subsection{Training and Test Time of Different Approaches}
\label{ssc:ta}
In this section, we compared the training and test time of ROCKET, ITime, LSTime, KDTime, and KDCTIme, ignoring KDCT or KDCR, as their training and inference time are of no significant difference. Instead of showing all the results in a table, diagrams of comparison were selected for better illustrating the results.

\begin{figure}[!t]
	\centering
	\subfigure[KDCTime and ROCKET]{
		\includegraphics[width=0.45\columnwidth]{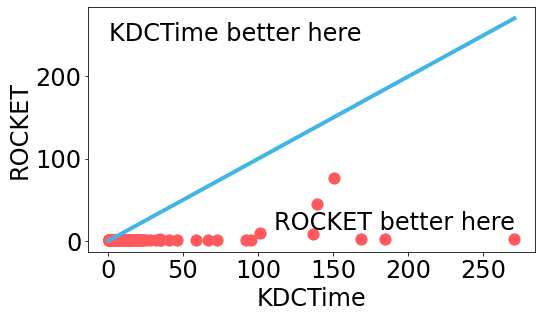}
	}
	\hfil
	\subfigure[KDCTime and ITime]{
		\includegraphics[width=0.45\columnwidth]{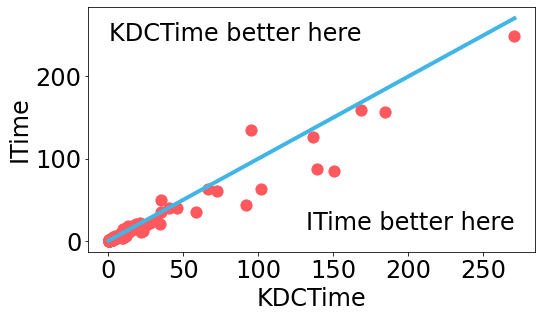}
	}
	\subfigure[KDCTime and LSTime]{
		\includegraphics[width=0.45\columnwidth]{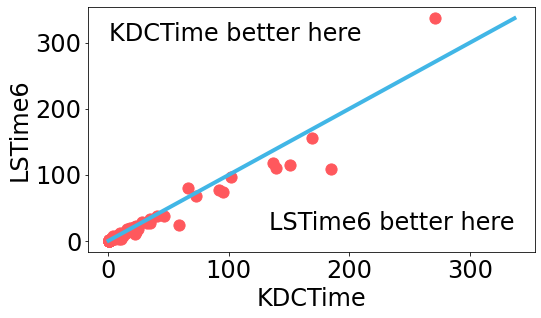}
	}
	\hfil
	\subfigure[KDCTime and KDTime]{
		\includegraphics[width=0.45\columnwidth]{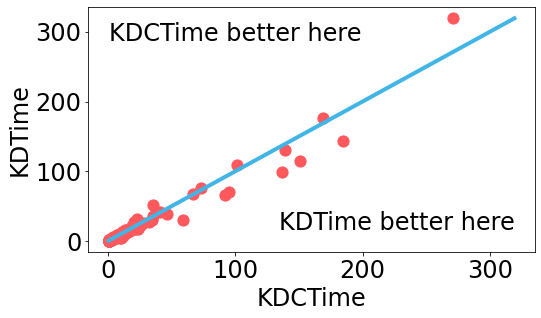}
	}
	\caption{The training time of KDCTime compared with ROCKET, ITime, LSTime, and KDTime}
	\label{fg:traint}
\end{figure}

Fig. \ref{fg:traint} demonstrates the training time of KDCTime compared with ROCKET, ITime, LSTime, and KDTime. In Fig. \ref{fg:traint}(a), it shows that KDCTime is much slower. In detail, the training time of KDCTime on $77$ datasets is below $1$ order of magnitude slower than ROCKET, while that on $36$ datasets is more than $1$ order of magnitude slower. That is because all ITime-based approaches, including KDCTime, employ Gradient Descent as the optimizing algorithm, which requires the inference on many training samples for each update and a large number of updates in total, e.g. $64$ samples for each update, many times of updating in $1$ epoch, and $512$ epochs in total. To the opposite, ROCKET utilizes ridge classifier and solves the ridge regression problem directly. However, the training time of KDCTime is still acceptable, which requires around $1$ hour to train $113$ UCR datasets and $10$ hours for $10$ times running in total. Besides, as shown in Fig. \ref{fg:traint}(b) and (c), the training time of ITime and LSTime is similar to KDCTime. The reason is that their model is the same, which is InceptionTime with $3$ Inception modules. Still, KDCTime needs an extra teacher model in order to guild the training of its student model, which leads to a consequence that KDCTime requires extra training time for the teacher model. Note that the training of the teacher model is required for only one time. Once the teacher model is obtained, that model is available for multiple times of training for the student model. Fig. \ref{fg:traint}(d) shows that the training time of KDCTime is of no significant difference with KDTime, as their models are the same, they both require the teacher model, and Gradient Descent is selected as their optimizing algorithms. As a conclusion, the difference of training time mainly appears in $3$ categories of methods, where the first one is ROCKET, the second one is ITime-based methods without KD, and the last one is ITime-based approaches with KD.

\begin{figure}[!t]
	\centering
	\subfigure[KDCTime and ROCKET]{
		\includegraphics[width=0.45\columnwidth]{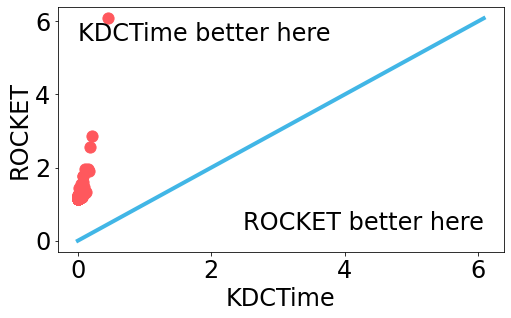}
	}
	\hfil
	\subfigure[KDCTime and ITime]{
		\includegraphics[width=0.45\columnwidth]{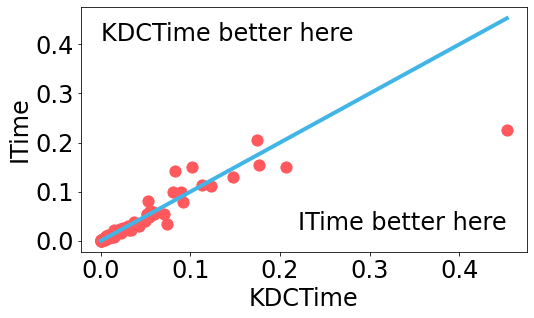}
	}
	\subfigure[KDCTime and LSTime]{
		\includegraphics[width=0.45\columnwidth]{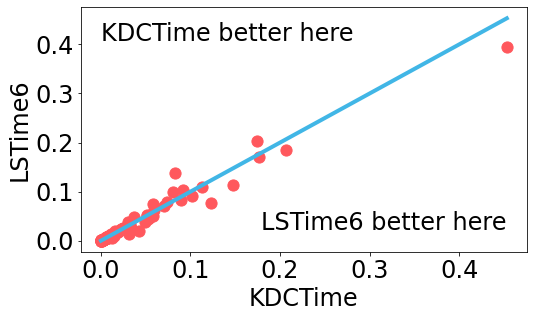}
	}
	\hfil
	\subfigure[KDCTime and KDTime]{
		\includegraphics[width=0.45\columnwidth]{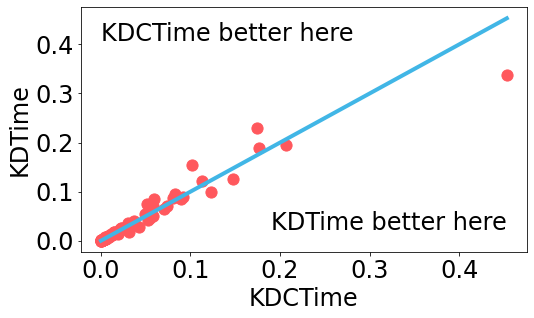}
	}
	\caption{The test time of KDCTime compared with ROCKET, ITime, LSTime, and KDTime}
	\label{fg:testt}
\end{figure}

Fig. \ref{fg:testt} demonstrates the test time of KDCTime compared with ROCKET, ITime, LSTime, and KDTime. In Fig. \ref{fg:testt}(a), it shows that KDCTime is much faster than ROCKET. To be specific, the test time of KDCTime on $42$ datasets is $1$ order of magnitude faster than ROCKET, that on $61$ datasets is $2$ orders of magnitude faster, and that on $10$ datasets is $3$ orders of magnitude faster. That is because in the stage of test, without the computational time of Gradient Descent, KDCTime only requires the time of inference on test samples, which is same as ROCKET. In this scenario, the computational time of $10000$ random convolutional kernels in ROCKET is much slower than the $3$ Inception modules in  KDCTime. In addition, as shown in Fig. \ref{fg:testt}(b), (c) and (d), the test time of those approaches are of no difference with KDCTime, since their inference models are all InceptionTime with $3$ Inception modules. As a consequence, the difference of test time can be categorized as $2$ groups, which are ROCKET and ITime-based approaches, regardless of ITime-based approaches with or without KD.

\section{Conclusion}
\label{sc:conc}
In this paper, we discovered the DNN-based TSC approaches are easy to be overfitting on the UCR datasets, which is caused by the few-shot problem in the UCR archive. Thus, in order to alleviate overfitting, Label Smoothing for InceptionTime (LSTime) was first proposed by utilizing soft labels. Next, instead of manually adjusting soft labels, Knowledge Distillation for InceptionTime (KDTime) was proposed in order to automatically generate soft labels. At last, in order to rectify the incorrect predicted soft labels from the teacher model, KD with calibration (KDC) was proposed, where it has two optional strategies, namely KDC by Translating (KDCT) and KDC by Reordering (KDCR).

The experimental results show that the accuracy of KDCT and KDCR is promising, while KDCR gets the highest one. In addition, including KDCT and KDCR, all InceptionTime-based (ITime-based) approaches are $2$ orders of magnitude faster than ROCKET on test time, since the ITime model is the majority factor for the inference time. The training time of ITime-based approaches are slower than ROCKET, yet it is in an acceptable range and worthwhile in order to obtain a promising accuracy and fast inference time. At last, KDCT and KDCR do not introduce any additional hyperparameter compared to ITime.

In the future, instead of just concentrating on the loss functions and labels, we will try various models, in order to propose a brand new model which owns a high generalization capability.

\appendices
\section{The procedure to calculate $\delta$}
\label{ap:pcd}
Given a hard label $\mathbf{y}^{h}$ and a soft label $\mathbf{y}^{t}$, where they both satisfy $\sum_{i=1}^{C}y_{i}=1$ and $\forall i, y_{i}\geq0$. By treating $\mathbf{y}^{h}$ and $\mathbf{y}^{t}$ as vectors, we want to find the minimum distance between them when $\mathop{\arg\max}_{i}\{y^{t}_{i}\}\neq\mathop{\arg\max}_{i}\{y^{h}_{i}\}$. Let $c=\mathop{\arg\max}_{i}\{y^{h}_{i}\}$, so that $y^{h}_{c}=1$ and $c\neq\mathop{\arg\max}_{i}\{y^{t}_{i}\}$. Let $m=\mathop{\arg\max}_{i}\{y^{t}_{i}\}$, so that $\forall i,y^{t}_{i}\leq y^{t}_{m}$. Therefore, we have the following optimization objective:

\begin{equation*}
	\label{eq:doo}
	\begin{aligned}
		\min&\Vert\mathbf{y}^{h}-\mathbf{y}^{t}\Vert_{2}\\
		\mathrm{s.t.}&\sum_{i=1}^{C}y^{t}_{i}=1\\
		&y^{t}_{i}\geq0, i=1,2,\dots,C\\
		&y^{t}_{i}\leq y^{t}_{m}, i=1,2,\dots,C
	\end{aligned}
\end{equation*}
where we know $y^{h}_{c}=1$ and $\forall i\neq c,y^{h}_{i}=0$. Thus, $\min\Vert\mathbf{y}^{h}-\mathbf{y}^{t}\Vert_{2}=\min\{(1-y^{t}_{c})^{2}+\sum_{i\neq c}(y^{t}_{i})^{2}\}$. Since $\sum_{i=1}^{C}y^{t}_{i}=1$, we let $y^{t}_{c}=1-\sum_{i\neq c}y^{t}_{i}$. Thus, $\min\{(1-y^{t}_{c})^{2}+\sum_{i\neq c}(y^{t}_{i})^{2}\}=\min\{(\sum_{i\neq c}y^{t}_{i})^{2}+\sum_{i\neq c}(y^{t}_{i})^{2}\}$. In this way, our optimization objective can be rewritten as follows:

\begin{equation*}
	\label{eq:dood}
	\begin{aligned}
		\min&\{(\sum_{i\neq c}y^{t}_{i})^{2}+\sum_{i\neq c}(y^{t}_{i})^{2}\}\\
		\mathrm{s.t.}&-y^{t}_{i}\leq0,i=1,2,\dots,C\\
		&y^{t}_{i}-y^{t}_{m}\leq0,i=1,2,\dots,C
	\end{aligned}
\end{equation*}

This is an optimization problem with inequality constraints. Therefore, we can define its Lagrangian function as:

\begin{equation*}
\label{eq:lag}
\begin{aligned}
	&\mathcal{L}(\mathbf{y}^{t},\boldsymbol{\lambda},\boldsymbol{\mu})\\
	&=(\sum_{i\neq c}y^{t}_{i})^{2}+\sum_{i\neq c}(y^{t}_{i})^{2}-\sum_{i }\lambda_{i}y^{t}_{i}+\sum_{i }\mu_{i}(y^{t}_{i}-y^{t}_{m})
\end{aligned}
\end{equation*}
where $\boldsymbol{\lambda}\in \mathbb{R}^{C}$ and $\boldsymbol{\mu}\in \mathbb{R}^{C}$ are two sets of Lagrangian multipliers. By adopting Karush-Kuhn-Tucker (KKT) Conditions, we have:

\begin{equation*}
\label{eq:kkt}
	\begin{cases}
		-\lambda_{c}+\mu_{c}=0,i=c\\
		2\sum_{j\neq c}y^{t}_{j}+2y^{t}_{m}-\lambda_{m}-\sum_{j\neq m}\mu_{j}=0,i=m\\
		2\sum_{j\neq c}y^{t}_{j}+2y^{t}_{i}-\lambda_{i}+\mu_{i}=0,i\neq c~and~i\neq m\\
		-\lambda_{i}y^{t}_{i}=0,i=0,1,\dots,C\\
		\mu_{i}(y^{t}_{i}-y^{t}_{m})=0,i=0,1,\dots,C\\
		\lambda_{i}\geq0,i=0,1,\dots,C\\
		\mu_{i}\geq0,i=0,1,\dots,C
	\end{cases}
\end{equation*}

At last, after solving this system of equations, we know $\mathcal{L}(\mathbf{y}^{t},\boldsymbol{\lambda},\boldsymbol{\mu})$ obtains the minimum value when $y^{t}_{c}=1/2$, $y^{t}_{m}=1/2$, and all other $y^{t}_{i}=0$. As a result, $\delta$ can be calculated as follows:

\begin{equation*}
\label{eq:delta}
	\delta=\min\Vert\mathbf{y}^{h}-\mathbf{y}^{t}\Vert_{2}=\sqrt{(1-\frac{1}{2})^{2}+(-\frac{1}{2})^2}=\frac{1}{\sqrt{2}}
\end{equation*}

\section{The accuracy of different approaches}
\label{ap:aa}
The accuracy of different approaches on UCR datasets is given in Table \ref{tb:aa} on the last page.

\begin{table*}[!t]\tiny
	\centering
	\caption{The accuracy of different approaches}
	\label{tb:aa}
	\begin{tabular}{ccccccc}
		\hline
		\textbf{Datasets} & \textbf{ROCKET} & \textbf{ITime} & \textbf{LSTime} & \textbf{KDTime} & \textbf{KDCT} & \textbf{KDCR}\\
		\hline
		ACSF1 & $0.882\pm0.01$ & $0.896\pm0.007$ & $0.903\pm\mathbf{0.004}$ & $0.911\pm0.021$ & $\mathbf{0.914}\pm0.006$ & $0.913\pm0.009$ \\
		Adiac & $0.8\pm\mathbf{0.005}$ & $0.822\pm0.02$ & $0.791\pm0.007$ & $0.842\pm0.007$ & $\mathbf{0.847}\pm0.009$ & $0.826\pm0.014$ \\
		ArrowHead & $0.807\pm\mathbf{0.009}$ & $0.84\pm0.018$ & $0.854\pm0.012$ & $0.857\pm0.016$ & $0.859\pm0.025$ & $\mathbf{0.873}\pm0.019$ \\
		Beef & $\mathbf{0.8}\pm\mathbf{0.0}$ & $0.767\pm\mathbf{0.0}$ & $0.767\pm\mathbf{0.0}$ & $0.793\pm0.057$ & $0.793\pm0.025$ & $\mathbf{0.8}\pm0.047$ \\
		BeetleFly & $0.9\pm\mathbf{0.0}$ & $0.74\pm0.02$ & $0.8\pm0.041$ & $0.87\pm0.024$ & $\mathbf{0.94}\pm0.02$ & $0.83\pm0.06$ \\
		BirdChicken & $0.89\pm\mathbf{0.02}$ & $0.92\pm0.024$ & $0.917\pm0.085$ & $\mathbf{0.95}\pm0.055$ & $0.92\pm0.024$ & $\mathbf{0.95}\pm0.063$ \\
		BME & $\mathbf{1.0}\pm\mathbf{0.0}$ & $0.995\pm\mathbf{0.0}$ & $0.995\pm0.004$ & $0.999\pm0.002$ & $0.992\pm0.003$ & $\mathbf{1.0}\pm0.002$ \\
		Car & $0.917\pm\mathbf{0.0}$ & $0.91\pm0.017$ & $0.867\pm0.014$ & $\mathbf{0.937}\pm0.012$ & $0.91\pm0.023$ & $0.917\pm0.008$ \\
		CBF & $\mathbf{1.0}\pm\mathbf{0.0}$ & $0.998\pm0.001$ & $0.998\pm0.001$ & $0.999\pm\mathbf{0.0}$ & $0.999\pm\mathbf{0.0}$ & $\mathbf{1.0}\pm\mathbf{0.0}$ \\
		Chinatown & $0.977\pm\mathbf{0.0}$ & $0.984\pm0.005$ & $\mathbf{0.992}\pm\mathbf{0.0}$ & $\mathbf{0.992}\pm0.001$ & $0.991\pm0.002$ & $\mathbf{0.992}\pm\mathbf{0.0}$ \\
		ChlorineConcentration & $0.81\pm0.006$ & $0.839\pm0.007$ & $0.83\pm0.016$ & $0.864\pm\mathbf{0.004}$ & $0.848\pm0.01$ & $\mathbf{0.873}\pm0.005$ \\
		CinCECGTorso & $0.826\pm\mathbf{0.005}$ & $0.863\pm0.009$ & $0.846\pm0.007$ & $0.868\pm0.012$ & $0.871\pm0.009$ & $\mathbf{0.872}\pm0.008$ \\
		Coffee & $\mathbf{1.0}\pm\mathbf{0.0}$ & $\mathbf{1.0}\pm\mathbf{0.0}$ & $\mathbf{1.0}\pm\mathbf{0.0}$ & $\mathbf{1.0}\pm\mathbf{0.0}$ & $\mathbf{1.0}\pm\mathbf{0.0}$ & $\mathbf{1.0}\pm\mathbf{0.0}$ \\
		Computers & $0.761\pm0.005$ & $0.813\pm0.014$ & $0.831\pm0.005$ & $0.826\pm\mathbf{0.003}$ & $\mathbf{0.851}\pm0.014$ & $0.818\pm0.01$ \\
		CricketX & $0.829\pm0.004$ & $0.855\pm0.014$ & $0.852\pm0.005$ & $0.86\pm\mathbf{0.002}$ & $\mathbf{0.871}\pm0.009$ & $0.85\pm0.008$ \\
		CricketY & $0.856\pm\mathbf{0.002}$ & $0.849\pm0.017$ & $0.873\pm0.008$ & $0.875\pm0.006$ & $0.879\pm0.008$ & $\mathbf{0.893}\pm0.009$ \\
		CricketZ & $0.858\pm\mathbf{0.003}$ & $0.856\pm0.007$ & $0.858\pm\mathbf{0.003}$ & $\mathbf{0.867}\pm0.006$ & $0.865\pm0.008$ & $0.865\pm0.007$ \\
		Crop & $0.763\pm\mathbf{0.001}$ & $0.77\pm0.002$ & $0.776\pm0.003$ & $0.777\pm0.002$ & $0.776\pm\mathbf{0.001}$ & $\mathbf{0.788}\pm\mathbf{0.001}$ \\
		DiatomSizeReduction & $\mathbf{0.975}\pm\mathbf{0.001}$ & $0.93\pm0.005$ & $0.938\pm0.023$ & $0.965\pm0.013$ & $0.942\pm0.013$ & $0.951\pm0.003$ \\
		DistalPhalanxOutlineAgeGroup & $0.758\pm\mathbf{0.006}$ & $0.795\pm0.011$ & $\mathbf{0.845}\pm0.01$ & $0.844\pm0.012$ & $0.836\pm0.007$ & $0.818\pm0.01$ \\
		DistalPhalanxOutlineCorrect & $0.773\pm\mathbf{0.005}$ & $0.762\pm0.017$ & $0.815\pm0.009$ & $0.816\pm0.011$ & $\mathbf{0.818}\pm0.011$ & $0.808\pm0.007$ \\
		DistalPhalanxTW & $\mathbf{0.717}\pm0.007$ & $0.626\pm0.041$ & $\mathbf{0.717}\pm\mathbf{0.004}$ & $0.635\pm0.018$ & $0.711\pm0.016$ & $0.713\pm0.011$ \\
		Earthquakes & $\mathbf{0.748}\pm\mathbf{0.0}$ & $0.647\pm0.038$ & $0.724\pm0.006$ & $0.736\pm0.013$ & $0.742\pm0.013$ & $0.742\pm0.012$ \\
		ECG200 & $0.906\pm\mathbf{0.005}$ & $0.887\pm0.013$ & $0.905\pm0.009$ & $0.915\pm0.01$ & $0.921\pm\mathbf{0.005}$ & $\mathbf{0.931}\pm0.007$ \\
		ECG5000 & $\mathbf{0.947}\pm\mathbf{0.0}$ & $0.94\pm\mathbf{0.0}$ & $0.946\pm\mathbf{0.0}$ & $0.945\pm0.001$ & $\mathbf{0.947}\pm\mathbf{0.0}$ & $0.944\pm0.001$ \\
		ECGFiveDays & $\mathbf{1.0}\pm\mathbf{0.0}$ & $0.998\pm0.001$ & $0.999\pm0.001$ & $\mathbf{1.0}\pm0.001$ & $0.999\pm0.001$ & $\mathbf{1.0}\pm\mathbf{0.0}$ \\
		ElectricDevices & $0.73\pm\mathbf{0.002}$ & $0.697\pm0.006$ & $0.733\pm0.008$ & $0.741\pm0.004$ & $0.738\pm0.006$ & $\mathbf{0.756}\pm0.004$ \\
		EOGHorizontalSignal & $0.578\pm\mathbf{0.004}$ & $0.661\pm0.005$ & $0.651\pm0.007$ & $0.691\pm0.01$ & $0.693\pm0.005$ & $\mathbf{0.713}\pm0.007$ \\
		EOGVerticalSignal & $0.547\pm0.006$ & $0.496\pm0.014$ & $0.497\pm\mathbf{0.005}$ & $0.541\pm0.017$ & $0.542\pm0.011$ & $\mathbf{0.563}\pm0.009$ \\
		EthanolLevel & $0.588\pm0.008$ & $0.783\pm0.017$ & $0.705\pm0.017$ & $0.762\pm\mathbf{0.005}$ & $\mathbf{0.817}\pm0.01$ & $0.816\pm0.02$ \\
		FaceAll & $0.923\pm\mathbf{0.007}$ & $0.816\pm0.01$ & $0.833\pm0.009$ & $0.897\pm0.028$ & $0.912\pm0.026$ & $\mathbf{0.963}\pm0.022$ \\
		FaceFour & $\mathbf{0.975}\pm0.005$ & $0.956\pm\mathbf{0.0}$ & $0.956\pm\mathbf{0.0}$ & $0.96\pm0.008$ & $0.954\pm0.003$ & $0.956\pm0.008$ \\
		FacesUCR & $0.962\pm\mathbf{0.001}$ & $0.965\pm0.002$ & $0.964\pm0.002$ & $0.968\pm\mathbf{0.001}$ & $0.968\pm0.002$ & $\mathbf{0.973}\pm\mathbf{0.001}$ \\
		FiftyWords & $\mathbf{0.835}\pm\mathbf{0.001}$ & $0.729\pm0.003$ & $0.63\pm0.006$ & $0.677\pm0.022$ & $0.753\pm0.015$ & $0.755\pm0.014$ \\
		Fish & $0.983\pm\mathbf{0.0}$ & $0.984\pm0.003$ & $0.99\pm0.004$ & $\mathbf{0.992}\pm0.004$ & $0.99\pm0.003$ & $0.99\pm0.003$ \\
		FordA & $0.944\pm0.001$ & $0.948\pm0.002$ & $0.958\pm\mathbf{0.0}$ & $\mathbf{0.963}\pm0.001$ & $0.961\pm0.001$ & $0.962\pm0.001$ \\
		FordB & $0.803\pm0.005$ & $0.838\pm0.003$ & $0.856\pm0.003$ & $0.859\pm0.004$ & $0.861\pm\mathbf{0.002}$ & $\mathbf{0.863}\pm0.004$ \\
		FreezerRegularTrain & $0.998\pm\mathbf{0.0}$ & $0.997\pm\mathbf{0.0}$ & $\mathbf{0.999}\pm\mathbf{0.0}$ & $\mathbf{0.999}\pm\mathbf{0.0}$ & $\mathbf{0.999}\pm\mathbf{0.0}$ & $0.998\pm\mathbf{0.0}$ \\
		FreezerSmallTrain & $\mathbf{0.952}\pm\mathbf{0.004}$ & $0.872\pm\mathbf{0.004}$ & $0.874\pm0.008$ & $0.937\pm0.006$ & $0.935\pm0.008$ & $0.944\pm0.007$ \\
		Fungi & $0.987\pm0.003$ & $0.999\pm\mathbf{0.002}$ & $0.995\pm0.004$ & $0.997\pm0.003$ & $0.997\pm0.003$ & $\mathbf{1.0}\pm0.005$ \\
		GunPoint & $\mathbf{1.0}\pm\mathbf{0.0}$ & $\mathbf{1.0}\pm\mathbf{0.0}$ & $0.998\pm0.002$ & $0.999\pm0.002$ & $\mathbf{1.0}\pm\mathbf{0.0}$ & $\mathbf{1.0}\pm0.006$ \\
		GunPointAgeSpan & $0.992\pm0.003$ & $0.985\pm0.007$ & $0.992\pm\mathbf{0.001}$ & $0.992\pm0.005$ & $0.992\pm0.004$ & $\mathbf{0.994}\pm0.009$ \\
		GunPointMaleVersusFemale & $0.999\pm0.002$ & $\mathbf{1.0}\pm\mathbf{0.0}$ & $\mathbf{1.0}\pm\mathbf{0.0}$ & $\mathbf{1.0}\pm\mathbf{0.0}$ & $\mathbf{1.0}\pm\mathbf{0.0}$ & $\mathbf{1.0}\pm0.002$ \\
		GunPointOldVersusYoung & $\mathbf{1.0}\pm\mathbf{0.0}$ & $\mathbf{1.0}\pm\mathbf{0.0}$ & $\mathbf{1.0}\pm\mathbf{0.0}$ & $\mathbf{1.0}\pm\mathbf{0.0}$ & $\mathbf{1.0}\pm\mathbf{0.0}$ & $\mathbf{1.0}\pm\mathbf{0.0}$ \\
		Ham & $0.722\pm0.02$ & $0.767\pm0.02$ & $0.816\pm\mathbf{0.009}$ & $\mathbf{0.821}\pm0.012$ & $0.797\pm0.011$ & $0.818\pm0.029$ \\
		HandOutlines & $0.944\pm\mathbf{0.002}$ & $0.94\pm0.005$ & $0.911\pm0.011$ & $0.945\pm0.009$ & $0.95\pm0.006$ & $\mathbf{0.954}\pm0.01$ \\
		Haptics & $0.523\pm\mathbf{0.0}$ & $0.552\pm0.014$ & $0.576\pm0.012$ & $0.577\pm0.006$ & $\mathbf{0.579}\pm0.013$ & $0.562\pm0.007$ \\
		Herring & $0.691\pm\mathbf{0.012}$ & $0.725\pm0.035$ & $\mathbf{0.75}\pm0.013$ & $0.728\pm0.016$ & $0.707\pm0.023$ & $0.743\pm0.025$ \\
		HouseTwenty & $\mathbf{0.965}\pm\mathbf{0.003}$ & $0.934\pm\mathbf{0.003}$ & $0.948\pm0.004$ & $0.953\pm0.012$ & $0.943\pm0.007$ & $0.944\pm0.016$ \\
		InlineSkate & $0.456\pm\mathbf{0.004}$ & $0.446\pm0.015$ & $0.418\pm0.007$ & $\mathbf{0.474}\pm0.015$ & $0.471\pm0.031$ & $0.458\pm0.008$ \\
		InsectEPGRegularTrain & $\mathbf{1.0}\pm\mathbf{0.0}$ & $\mathbf{1.0}\pm\mathbf{0.0}$ & $\mathbf{1.0}\pm\mathbf{0.0}$ & $\mathbf{1.0}\pm\mathbf{0.0}$ & $\mathbf{1.0}\pm\mathbf{0.0}$ & $\mathbf{1.0}\pm\mathbf{0.0}$ \\
		InsectEPGSmallTrain & $0.998\pm0.002$ & $\mathbf{1.0}\pm\mathbf{0.0}$ & $\mathbf{1.0}\pm\mathbf{0.0}$ & $\mathbf{1.0}\pm\mathbf{0.0}$ & $\mathbf{1.0}\pm\mathbf{0.0}$ & $\mathbf{1.0}\pm\mathbf{0.0}$ \\
		InsectWingbeatSound & $\mathbf{0.652}\pm\mathbf{0.001}$ & $0.62\pm0.009$ & $0.635\pm0.005$ & $0.635\pm0.007$ & $0.631\pm0.007$ & $0.64\pm0.004$ \\
		ItalyPowerDemand & $0.969\pm\mathbf{0.0}$ & $0.972\pm0.002$ & $0.973\pm0.002$ & $0.974\pm0.001$ & $0.974\pm0.001$ & $\mathbf{0.975}\pm0.001$ \\
		LargeKitchenAppliances & $0.888\pm0.003$ & $0.89\pm0.008$ & $0.891\pm\mathbf{0.002}$ & $0.91\pm0.004$ & $0.897\pm0.005$ & $\mathbf{0.914}\pm0.006$ \\
		Lightning2 & $0.77\pm\mathbf{0.0}$ & $0.813\pm0.017$ & $0.831\pm0.008$ & $0.856\pm0.019$ & $0.859\pm0.017$ & $\mathbf{0.869}\pm0.028$ \\
		Lightning7 & $0.841\pm0.011$ & $0.822\pm0.027$ & $0.906\pm0.011$ & $0.905\pm\mathbf{0.009}$ & $0.916\pm0.012$ & $\mathbf{0.938}\pm0.046$ \\
		Mallat & $0.955\pm\mathbf{0.001}$ & $0.95\pm0.006$ & $0.834\pm0.011$ & $0.947\pm0.006$ & $0.945\pm0.007$ & $\mathbf{0.963}\pm0.014$ \\
		Meat & $0.95\pm\mathbf{0.0}$ & $0.913\pm0.019$ & $0.928\pm0.031$ & $0.913\pm0.037$ & $0.917\pm0.028$ & $\mathbf{0.967}\pm0.012$ \\
		MedicalImages & $\mathbf{0.802}\pm\mathbf{0.003}$ & $0.761\pm0.008$ & $0.794\pm0.007$ & $0.786\pm\mathbf{0.003}$ & $0.79\pm0.005$ & $0.779\pm0.004$ \\
		MiddlePhalanxOutlineAgeGroup & $0.603\pm\mathbf{0.003}$ & $0.562\pm0.01$ & $0.708\pm0.004$ & $0.582\pm\mathbf{0.003}$ & $\mathbf{0.71}\pm0.006$ & $0.706\pm0.019$ \\
		MiddlePhalanxOutlineCorrect & $0.839\pm0.007$ & $0.824\pm0.011$ & $0.853\pm0.004$ & $\mathbf{0.858}\pm0.007$ & $0.855\pm\mathbf{0.002}$ & $0.854\pm0.006$ \\
		MiddlePhalanxTW & $0.552\pm0.006$ & $0.498\pm0.026$ & $0.61\pm\mathbf{0.002}$ & $0.617\pm0.006$ & $\mathbf{0.62}\pm0.009$ & $0.598\pm0.03$ \\
		MixedShapesRegularTrain & $0.971\pm\mathbf{0.001}$ & $0.975\pm0.002$ & $\mathbf{0.976}\pm\mathbf{0.001}$ & $\mathbf{0.976}\pm\mathbf{0.001}$ & $0.974\pm\mathbf{0.001}$ & $0.975\pm0.002$ \\
		MixedShapesSmallTrain & $\mathbf{0.937}\pm\mathbf{0.0}$ & $0.932\pm0.003$ & $0.932\pm0.004$ & $\mathbf{0.937}\pm0.001$ & $0.935\pm0.006$ & $0.935\pm0.006$ \\
		MoteStrain & $\mathbf{0.91}\pm\mathbf{0.002}$ & $0.9\pm0.008$ & $0.908\pm0.008$ & $0.896\pm0.005$ & $0.908\pm0.004$ & $0.874\pm0.026$ \\
		NonInvasiveFetalECGThorax1 & $0.954\pm0.002$ & $0.958\pm0.003$ & $0.952\pm0.005$ & $0.961\pm0.003$ & $\mathbf{0.964}\pm0.003$ & $\mathbf{0.964}\pm\mathbf{0.001}$ \\
		NonInvasiveFetalECGThorax2 & $\mathbf{0.97}\pm\mathbf{0.001}$ & $0.959\pm0.004$ & $0.96\pm0.002$ & $0.961\pm0.003$ & $0.961\pm\mathbf{0.001}$ & $0.961\pm0.002$ \\
		OliveOil & $\mathbf{0.907}\pm\mathbf{0.013}$ & $0.753\pm0.016$ & $0.722\pm0.079$ & $0.633\pm0.191$ & $0.64\pm0.196$ & $0.767\pm0.033$ \\
		OSULeaf & $0.939\pm\mathbf{0.004}$ & $0.97\pm0.007$ & $0.97\pm\mathbf{0.004}$ & $\mathbf{0.981}\pm\mathbf{0.004}$ & $0.978\pm0.005$ & $0.972\pm0.006$ \\
		PhalangesOutlinesCorrect & $0.836\pm0.003$ & $0.831\pm0.008$ & $\mathbf{0.855}\pm\mathbf{0.002}$ & $\mathbf{0.855}\pm0.008$ & $0.848\pm0.003$ & $0.851\pm0.003$ \\
		Phoneme & $0.281\pm\mathbf{0.002}$ & $0.339\pm0.007$ & $0.344\pm0.006$ & $\mathbf{0.346}\pm0.004$ & $\mathbf{0.346}\pm0.003$ & $0.331\pm0.003$ \\
		PigAirwayPressure & $\mathbf{0.834}\pm0.028$ & $0.577\pm0.018$ & $0.367\pm0.032$ & $0.418\pm0.034$ & $0.592\pm0.02$ & $0.57\pm\mathbf{0.008}$ \\
		PigArtPressure & $0.962\pm0.004$ & $0.995\pm0.002$ & $0.863\pm0.006$ & $\mathbf{0.996}\pm\mathbf{0.0}$ & $\mathbf{0.996}\pm\mathbf{0.0}$ & $0.992\pm0.014$ \\
		PigCVP & $\mathbf{0.928}\pm\mathbf{0.0}$ & $0.77\pm0.073$ & $0.677\pm0.057$ & $0.77\pm0.018$ & $0.854\pm0.022$ & $0.866\pm0.025$ \\
		Plane & $\mathbf{1.0}\pm\mathbf{0.0}$ & $\mathbf{1.0}\pm\mathbf{0.0}$ & $\mathbf{1.0}\pm\mathbf{0.0}$ & $\mathbf{1.0}\pm\mathbf{0.0}$ & $\mathbf{1.0}\pm\mathbf{0.0}$ & $\mathbf{1.0}\pm0.008$ \\
		PowerCons & $0.979\pm0.005$ & $0.984\pm0.006$ & $0.997\pm0.002$ & $\mathbf{1.0}\pm\mathbf{0.0}$ & $0.994\pm0.004$ & $\mathbf{1.0}\pm\mathbf{0.0}$ \\
		ProximalPhalanxOutlineAgeGroup & $0.858\pm\mathbf{0.002}$ & $0.842\pm0.007$ & $0.866\pm\mathbf{0.002}$ & $\mathbf{0.872}\pm\mathbf{0.002}$ & $0.869\pm0.005$ & $0.868\pm0.003$ \\
		ProximalPhalanxOutlineCorrect & $0.897\pm0.007$ & $0.895\pm0.011$ & $0.925\pm0.009$ & $0.918\pm\mathbf{0.006}$ & $0.925\pm0.008$ & $\mathbf{0.927}\pm0.007$ \\
		ProximalPhalanxTW & $0.814\pm\mathbf{0.004}$ & $0.778\pm0.018$ & $\mathbf{0.863}\pm0.006$ & $0.862\pm0.006$ & $0.861\pm0.005$ & $0.852\pm0.016$ \\
		RefrigerationDevices & $0.533\pm0.01$ & $0.498\pm0.009$ & $0.556\pm0.014$ & $0.565\pm0.008$ & $\mathbf{0.566}\pm\mathbf{0.007}$ & $0.529\pm0.01$ \\
		Rock & $0.68\pm\mathbf{0.0}$ & $0.744\pm0.039$ & $0.62\pm0.028$ & $0.704\pm0.067$ & $0.733\pm0.037$ & $\mathbf{0.768}\pm0.029$ \\
		ScreenType & $0.477\pm0.014$ & $0.604\pm0.008$ & $0.633\pm\mathbf{0.003}$ & $0.631\pm0.006$ & $0.635\pm0.007$ & $\mathbf{0.65}\pm0.011$ \\
		SemgHandGenderCh2 & $\mathbf{0.901}\pm\mathbf{0.005}$ & $0.878\pm0.009$ & $0.878\pm\mathbf{0.005}$ & $0.885\pm\mathbf{0.005}$ & $0.891\pm0.006$ & $0.88\pm0.006$ \\
		SemgHandMovementCh2 & $\mathbf{0.591}\pm0.006$ & $0.446\pm0.018$ & $0.451\pm\mathbf{0.002}$ & $0.462\pm0.012$ & $0.469\pm0.023$ & $0.461\pm0.03$ \\
		SemgHandSubjectCh2 & $\mathbf{0.843}\pm0.007$ & $0.652\pm0.025$ & $0.669\pm0.011$ & $0.693\pm\mathbf{0.006}$ & $0.695\pm0.016$ & $0.688\pm0.029$ \\
		ShapeletSim & $\mathbf{1.0}\pm\mathbf{0.0}$ & $0.712\pm0.03$ & $0.997\pm0.002$ & $0.967\pm0.041$ & $0.991\pm0.007$ & $\mathbf{1.0}\pm0.002$ \\
		ShapesAll & $0.91\pm\mathbf{0.001}$ & $0.911\pm0.003$ & $0.895\pm0.004$ & $0.916\pm0.002$ & $0.919\pm0.003$ & $\mathbf{0.922}\pm0.003$ \\
		SmallKitchenAppliances & $\mathbf{0.82}\pm0.006$ & $0.753\pm0.007$ & $0.814\pm\mathbf{0.004}$ & $0.817\pm0.007$ & $0.813\pm0.008$ & $0.803\pm0.008$ \\
		SmoothSubspace & $0.979\pm0.003$ & $0.974\pm0.006$ & $0.99\pm0.004$ & $0.993\pm0.004$ & $0.991\pm\mathbf{0.002}$ & $\mathbf{1.0}\pm0.008$ \\
		SonyAIBORobotSurface1 & $0.918\pm\mathbf{0.001}$ & $0.887\pm0.014$ & $0.916\pm0.026$ & $0.902\pm0.026$ & $0.903\pm0.028$ & $\mathbf{0.969}\pm0.027$ \\
		SonyAIBORobotSurface2 & $0.918\pm\mathbf{0.003}$ & $\mathbf{0.97}\pm0.006$ & $0.962\pm0.004$ & $0.961\pm\mathbf{0.003}$ & $0.964\pm\mathbf{0.003}$ & $0.953\pm0.012$ \\
		StarLightCurves & $\mathbf{0.981}\pm0.001$ & $0.977\pm0.001$ & $0.98\pm\mathbf{0.0}$ & $0.98\pm\mathbf{0.0}$ & $\mathbf{0.981}\pm\mathbf{0.0}$ & $\mathbf{0.981}\pm\mathbf{0.0}$ \\
		Strawberry & $0.981\pm0.002$ & $0.982\pm0.002$ & $0.983\pm0.002$ & $\mathbf{0.985}\pm\mathbf{0.001}$ & $0.983\pm0.002$ & $0.981\pm0.002$ \\
		SwedishLeaf & $0.964\pm\mathbf{0.002}$ & $0.968\pm0.005$ & $0.972\pm0.004$ & $0.974\pm0.004$ & $0.973\pm0.005$ & $\mathbf{0.975}\pm0.003$ \\
		Symbols & $0.974\pm\mathbf{0.001}$ & $0.981\pm0.003$ & $0.976\pm0.005$ & $0.983\pm\mathbf{0.001}$ & $\mathbf{0.984}\pm0.002$ & $0.975\pm0.004$ \\
		SyntheticControl & $0.999\pm0.002$ & $0.996\pm0.001$ & $0.998\pm0.001$ & $0.998\pm0.001$ & $0.997\pm\mathbf{0.0}$ & $\mathbf{1.0}\pm0.002$ \\
		ToeSegmentation1 & $0.968\pm\mathbf{0.002}$ & $0.973\pm0.006$ & $0.98\pm0.003$ & $0.97\pm0.006$ & $0.976\pm0.004$ & $\mathbf{0.988}\pm0.012$ \\
		ToeSegmentation2 & $0.942\pm0.006$ & $0.956\pm0.005$ & $0.962\pm\mathbf{0.002}$ & $0.967\pm0.01$ & $0.964\pm0.006$ & $\mathbf{0.969}\pm0.004$ \\
		Trace & $\mathbf{1.0}\pm\mathbf{0.0}$ & $\mathbf{1.0}\pm\mathbf{0.0}$ & $\mathbf{1.0}\pm\mathbf{0.0}$ & $\mathbf{1.0}\pm\mathbf{0.0}$ & $\mathbf{1.0}\pm\mathbf{0.0}$ & $\mathbf{1.0}\pm\mathbf{0.0}$ \\
		TwoLeadECG & $\mathbf{0.999}\pm\mathbf{0.0}$ & $0.997\pm0.002$ & $0.998\pm\mathbf{0.0}$ & $\mathbf{0.999}\pm\mathbf{0.0}$ & $0.998\pm0.001$ & $\mathbf{0.999}\pm\mathbf{0.0}$ \\
		TwoPatterns & $\mathbf{1.0}\pm\mathbf{0.0}$ & $\mathbf{1.0}\pm\mathbf{0.0}$ & $\mathbf{1.0}\pm\mathbf{0.0}$ & $\mathbf{1.0}\pm\mathbf{0.0}$ & $\mathbf{1.0}\pm\mathbf{0.0}$ & $\mathbf{1.0}\pm\mathbf{0.0}$ \\
		UMD & $0.986\pm\mathbf{0.0}$ & $\mathbf{0.995}\pm\mathbf{0.0}$ & $0.993\pm0.002$ & $\mathbf{0.995}\pm\mathbf{0.0}$ & $\mathbf{0.995}\pm\mathbf{0.0}$ & $\mathbf{0.995}\pm0.003$ \\
		UWaveGestureLibraryAll & $\mathbf{0.975}\pm\mathbf{0.001}$ & $0.894\pm0.002$ & $0.898\pm0.002$ & $0.897\pm0.002$ & $0.897\pm0.003$ & $0.899\pm0.002$ \\
		UWaveGestureLibraryX & $\mathbf{0.85}\pm\mathbf{0.001}$ & $0.805\pm0.004$ & $0.817\pm0.002$ & $0.816\pm0.002$ & $0.816\pm0.003$ & $0.819\pm0.002$ \\
		UWaveGestureLibraryY & $\mathbf{0.773}\pm\mathbf{0.002}$ & $0.728\pm0.005$ & $0.749\pm0.004$ & $0.741\pm0.003$ & $0.737\pm0.004$ & $0.744\pm0.005$ \\
		UWaveGestureLibraryZ & $\mathbf{0.792}\pm\mathbf{0.002}$ & $0.761\pm\mathbf{0.002}$ & $0.77\pm0.006$ & $0.767\pm0.003$ & $0.768\pm0.003$ & $0.77\pm0.003$ \\
		Wafer & $0.999\pm\mathbf{0.0}$ & $0.998\pm\mathbf{0.0}$ & $\mathbf{1.0}\pm\mathbf{0.0}$ & $0.999\pm\mathbf{0.0}$ & $0.999\pm\mathbf{0.0}$ & $0.998\pm0.001$ \\
		Wine & $0.804\pm\mathbf{0.022}$ & $0.774\pm0.025$ & $0.833\pm0.079$ & $0.811\pm0.156$ & $0.833\pm0.059$ & $\mathbf{0.926}\pm0.03$ \\
		WordSynonyms & $\mathbf{0.753}\pm\mathbf{0.002}$ & $0.692\pm0.01$ & $0.644\pm0.011$ & $0.707\pm0.005$ & $0.699\pm0.006$ & $0.707\pm0.008$ \\
		Worms & $0.743\pm\mathbf{0.01}$ & $0.754\pm0.03$ & $0.738\pm0.011$ & $0.81\pm0.02$ & $\mathbf{0.822}\pm0.018$ & $0.752\pm0.04$ \\
		WormsTwoClass & $0.79\pm0.015$ & $0.808\pm0.03$ & $0.891\pm0.011$ & $0.884\pm\mathbf{0.009}$ & $0.9\pm\mathbf{0.009}$ & $\mathbf{0.906}\pm0.029$ \\
		Yoga & $\mathbf{0.912}\pm0.005$ & $0.909\pm\mathbf{0.004}$ & $0.898\pm\mathbf{0.004}$ & $0.908\pm0.008$ & $0.909\pm0.005$ & $0.91\pm0.006$ \\
		\hline
		Total accuracy wins & $40$ & $11$ & $17$ & $30$ & $29$ & $54$ \\
		Total std wins & $76$ & $21$ & $35$ & $34$ & $27$ & $17$ \\
		\hline
	\end{tabular}
\end{table*}



\bibliographystyle{IEEEtran}
\bibliography{IEEEabrv,TNNLS}

\begin{thebibliography}{10}
\providecommand{\url}[1]{#1}
\csname url@samestyle\endcsname
\providecommand{\newblock}{\relax}
\providecommand{\bibinfo}[2]{#2}
\providecommand{\BIBentrySTDinterwordspacing}{\spaceskip=0pt\relax}
\providecommand{\BIBentryALTinterwordstretchfactor}{4}
\providecommand{\BIBentryALTinterwordspacing}{\spaceskip=\fontdimen2\font plus
\BIBentryALTinterwordstretchfactor\fontdimen3\font minus
  \fontdimen4\font\relax}
\providecommand{\BIBforeignlanguage}[2]{{%
\expandafter\ifx\csname l@#1\endcsname\relax
\typeout{** WARNING: IEEEtran.bst: No hyphenation pattern has been}%
\typeout{** loaded for the language `#1'. Using the pattern for}%
\typeout{** the default language instead.}%
\else
\language=\csname l@#1\endcsname
\fi
#2}}
\providecommand{\BIBdecl}{\relax}
\BIBdecl

\bibitem{Fawaz:2019}
H.~I. Fawaz, G.~Forestier, J.~Weber, L.~Idoumghar, and P.~Muller, ``Deep
  learning for time series classification: a review,'' \emph{Data Mining and
  Knowledge Discovery}, vol.~33, no.~4, pp. 917--963, 2019.

\bibitem{Hu:2020}
J.~Hu, L.~Shen, S.~Albanie, G.~Sun, and E.~Wu, ``Squeeze-and-excitation
  networks,'' \emph{{IEEE} Transactions on Pattern Analysis and Machine
  Intelligence}, vol.~42, no.~8, pp. 2011--2023, 2020.

\bibitem{He:2016}
K.~He, X.~Zhang, S.~Ren, and J.~Sun, ``Deep residual learning for image
  recognition,'' in \emph{{IEEE} Conference on Computer Vision and Pattern
  Recognition}, 2016, pp. 770--778.

\bibitem{Szegedy:2015}
C.~Szegedy, W.~Liu, Y.~Jia, P.~Sermanet, S.~E. Reed, D.~Anguelov, D.~Erhan,
  V.~Vanhoucke, and A.~Rabinovich, ``Going deeper with convolutions,'' in
  \emph{{IEEE} Conference on Computer Vision and Pattern Recognition}, 2015,
  pp. 1--9.

\bibitem{UCRArchive}
W.~V. Anthony~Bagnall, Jason~Lines and E.~Keogh, ``The uea \& ucr time series
  classification repository,'' 2018, uRL:
  http://www.timeseriesclassification.com/.

\bibitem{Wang:2017}
Z.~Wang, W.~Yan, and T.~Oates, ``Time series classification from scratch with
  deep neural networks: {A} strong baseline,'' in \emph{International Joint
  Conference on Neural Networks}, 2017, pp. 1578--1585.

\bibitem{Fawaz:2020}
H.~I. Fawaz, B.~Lucas, G.~Forestier, C.~Pelletier, D.~F. Schmidt, J.~Weber,
  G.~I. Webb, L.~Idoumghar, P.~Muller, and F.~Petitjean, ``Inceptiontime:
  Finding alexnet for time series classification,'' \emph{Data Mining and
  Knowledge Discovery}, vol.~34, no.~6, pp. 1936--1962, 2020.

\bibitem{LeCun:1998}
Y.~LeCun, L.~Bottou, Y.~Bengio, and P.~Haffner, ``Gradient-based learning
  applied to document recognition,'' \emph{Proceedings of the IEEE}, vol.~86,
  no.~11, pp. 2278--2324, 1998.

\bibitem{CIFAR10}
A.~Krizhevsky, ``Learning multiple layers of features from tiny images,'' 2009,
  uRL: https://www.cs.utoronto.ca/$\sim$kriz/cifar.html.

\bibitem{Deng:2009}
J.~Deng, W.~Dong, R.~Socher, L.~Li, K.~Li, and L.~Fei{-}Fei, ``Imagenet: {A}
  large-scale hierarchical image database,'' in \emph{{IEEE} Conference on
  Computer Vision and Pattern Recognition}, 2009, pp. 248--255.

\bibitem{Lines:2018}
J.~Lines, S.~Taylor, and A.~J. Bagnall, ``Time series classification with
  {HIVE-COTE:} the hierarchical vote collective of transformation-based
  ensembles,'' \emph{{ACM} Transactions on Knowledge Discovery from Data},
  vol.~12, no.~5, pp. 52:1--52:35, 2018.

\bibitem{Dempster:2020}
A.~Dempster, F.~Petitjean, and G.~I. Webb, ``{ROCKET:} exceptionally fast and
  accurate time series classification using random convolutional kernels,''
  \emph{Data Mining and Knowledge Discovery}, vol.~34, no.~5, pp. 1454--1495,
  2020.

\bibitem{Ioffe:2015}
S.~Ioffe and C.~Szegedy, ``Batch normalization: Accelerating deep network
  training by reducing internal covariate shift,'' in \emph{International
  Conference on Machine Learning}, 2015, pp. 448--456.

\bibitem{Srivastava:2014}
N.~Srivastava, G.~Hinton, A.~Krizhevsky, I.~Sutskever, and R.~Salakhutdinov,
  ``Dropout: a simple way to prevent neural networks from overfitting,''
  \emph{The Journal of Machine Learning Research}, vol.~15, no.~1, pp.
  1929--1958, 2014.

\bibitem{Szegedy:2016}
C.~Szegedy, V.~Vanhoucke, S.~Ioffe, J.~Shlens, and Z.~Wojna, ``Rethinking the
  inception architecture for computer vision,'' in \emph{IEEE Conference on
  Computer Vision and Pattern Recognition}, 2016, pp. 2818--2826.

\bibitem{Wan:2017}
Y.~Wan and Y.~Si, ``A formal approach to chart patterns classification in
  financial time series,'' \emph{Information Sciences}, vol. 411, pp. 151--175,
  2017.

\bibitem{Hinton:2015}
G.~Hinton, O.~Vinyals, and J.~Dean, ``Distilling the knowledge in a neural
  network,'' \emph{arXiv preprint arXiv:1503.02531}, 2015.

\bibitem{Bagnall:2017}
A.~J. Bagnall, J.~Lines, A.~Bostrom, J.~Large, and E.~J. Keogh, ``The great
  time series classification bake off: a review and experimental evaluation of
  recent algorithmic advances,'' \emph{Data Mining and Knowledge Discovery},
  vol.~31, no.~3, pp. 606--660, 2017.

\bibitem{Rakthanmanon:2012}
T.~Rakthanmanon, B.~J.~L. Campana, A.~Mueen, G.~E. A. P.~A. Batista, M.~B.
  Westover, Q.~Zhu, J.~Zakaria, and E.~J. Keogh, ``Searching and mining
  trillions of time series subsequences under dynamic time warping,'' in
  \emph{International Conference on Knowledge Discovery and Data Mining}, 2012,
  pp. 262--270.

\bibitem{Sakurai:2007}
Y.~Sakurai, C.~Faloutsos, and M.~Yamamuro, ``Stream monitoring under the time
  warping distance,'' in \emph{International Conference on Data Engineering},
  pp. 1046--1055.

\bibitem{Gong:2018}
X.~Gong, S.~Fong, and Y.~Si, ``Fast multi-subsequence monitoring on streaming
  time-series based on forward-propagation,'' \emph{Information Sciences}, vol.
  450, pp. 73--88, 2018.

\bibitem{Baydogan:2013}
M.~G. Baydogan, G.~C. Runger, and E.~Tuv, ``A bag-of-features framework to
  classify time series,'' \emph{{IEEE} Transactions on Pattern Analysis and
  Machine Intelligence}, vol.~35, no.~11, pp. 2796--2802, 2013.

\bibitem{Kate:2016}
R.~J. Kate, ``Using dynamic time warping distances as features for improved
  time series classification,'' \emph{Data Mining and Knowledge Discovery},
  vol.~30, no.~2, pp. 283--312, 2016.

\bibitem{Bagnall:2016}
A.~J. Bagnall, J.~Lines, J.~Hills, and A.~Bostrom, ``Time-series classification
  with {COTE:} the collective of transformation-based ensembles,'' in
  \emph{International Conference on Data Engineering}, 2016, pp. 1548--1549.

\bibitem{Cui:2016}
Z.~Cui, W.~Chen, and Y.~Chen, ``Multi-scale convolutional neural networks for
  time series classification,'' \emph{arXiv preprint arXiv:1603.06995}, 2016.

\bibitem{Karimi-Bidhendi:2018}
S.~Karimi{-}Bidhendi, F.~Munshi, and A.~Munshi, ``Scalable classification of
  univariate and multivariate time series,'' in \emph{International Conference
  on Big Data}, 2018, pp. 1598--1605.

\bibitem{Szegedy:2017}
C.~Szegedy, S.~Ioffe, V.~Vanhoucke, and A.~A. Alemi, ``Inception-v4,
  inception-resnet and the impact of residual connections on learning,'' in
  \emph{{AAAI} Conference on Artificial Intelligence}, 2017, pp. 4278--4284.

\bibitem{Chen:2021}
W.~Chen and K.~Shi, ``Multi-scale attention convolutional neural network for
  time series classification,'' \emph{Neural Networks}, vol. 136, pp. 126--140,
  2021.

\bibitem{Romero:2014}
A.~Romero, N.~Ballas, S.~E. Kahou, A.~Chassang, C.~Gatta, and Y.~Bengio,
  ``Fitnets: Hints for thin deep nets,'' in \emph{International Conference on
  Learning Representations}, 2015.

\bibitem{Yim:2017}
J.~Yim, D.~Joo, J.~Bae, and J.~Kim, ``A gift from knowledge distillation: Fast
  optimization, network minimization and transfer learning,'' in \emph{{IEEE}
  Conference on Computer Vision and Pattern Recognition}, 2017, pp. 7130--7138.

\bibitem{Svitov:2020}
D.~Svitov and S.~Alyamkin, ``Margindistillation: Distillation for margin-based
  softmax,'' \emph{arXiv preprint arXiv:2003.02586}, 2020.

\bibitem{Oki:2020}
H.~Oki, M.~Abe, J.~Miyao, and T.~Kurita, ``Triplet loss for knowledge
  distillation,'' in \emph{International Joint Conference on Neural Networks},
  2020, pp. 1--7.

\bibitem{Cho:2019}
J.~H. Cho and B.~Hariharan, ``On the efficacy of knowledge distillation,'' in
  \emph{International Conference on Computer Vision}, 2019, pp. 4793--4801.

\bibitem{Mirzadeh:2020}
S.~Mirzadeh, M.~Farajtabar, A.~Li, N.~Levine, A.~Matsukawa, and H.~Ghasemzadeh,
  ``Improved knowledge distillation via teacher assistant,'' in \emph{{AAAI}
  Conference on Artificial Intelligence}, 2020, pp. 5191--5198.

\bibitem{Friedman:1940}
M.~Friedman, ``A comparison of alternative tests of significance for the
  problem of m rankings,'' \emph{The Annals of Mathematical Statistics},
  vol.~11, no.~1, pp. 86--92, 1940.

\bibitem{Benavoli:2016}
A.~Benavoli, G.~Corani, and F.~Mangili, ``Should we really use post-hoc tests
  based on mean-ranks?'' \emph{The Journal of Machine Learning Research},
  vol.~17, no.~1, pp. 152--161, 2016.

\bibitem{Garcia:2008}
S.~Garcia and F.~Herrera, ``An extension on" statistical comparisons of
  classifiers over multiple data sets" for all pairwise comparisons.''
  \emph{Journal of machine learning research}, vol.~9, no.~12, 2008.

\bibitem{Demsar:2006}
J.~Dem{\v{s}}ar, ``Statistical comparisons of classifiers over multiple data
  sets,'' \emph{The Journal of Machine Learning Research}, vol.~7, pp. 1--30,
  2006.

\end{thebibliography}

\begin{IEEEbiography}[{\includegraphics[width=1in,height=1.25in,clip,keepaspectratio]{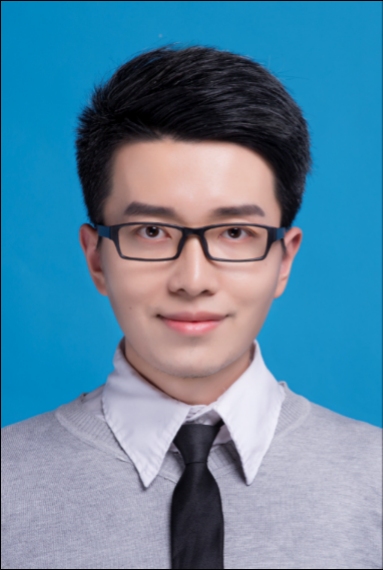}}]{Xueyuan Gong}
is currently an assistant professor at the school of Intelligent Systems Science and Engineering, Jinan University, Zhuhai, China. He obtained his Ph.D. and M.Sc. degree in Computer Science from University of Macau in 2014 and 2019 respectively. Before that, he received his B.Sc. degree from Macau University of Science and Technology in 2011. His research interests include Machine Learning and Data Mining.
\end{IEEEbiography}

%\newpage

\begin{IEEEbiography}[{\includegraphics[width=1in,height=1.25in,clip,keepaspectratio]{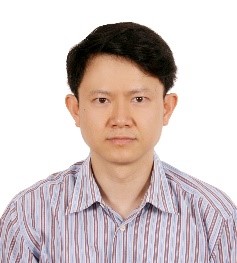}}]{Yain-Whar Si}
is an associate professor at the University of Macau. He holds a Ph.D. degree in Information Technology from the Queensland University of Technology, Australia. His research interests are in the areas of Financial Technology, Computational Intelligence, and Information Visualization.
\end{IEEEbiography}

\begin{IEEEbiography}[{\includegraphics[width=1in,height=1.25in,clip,keepaspectratio]{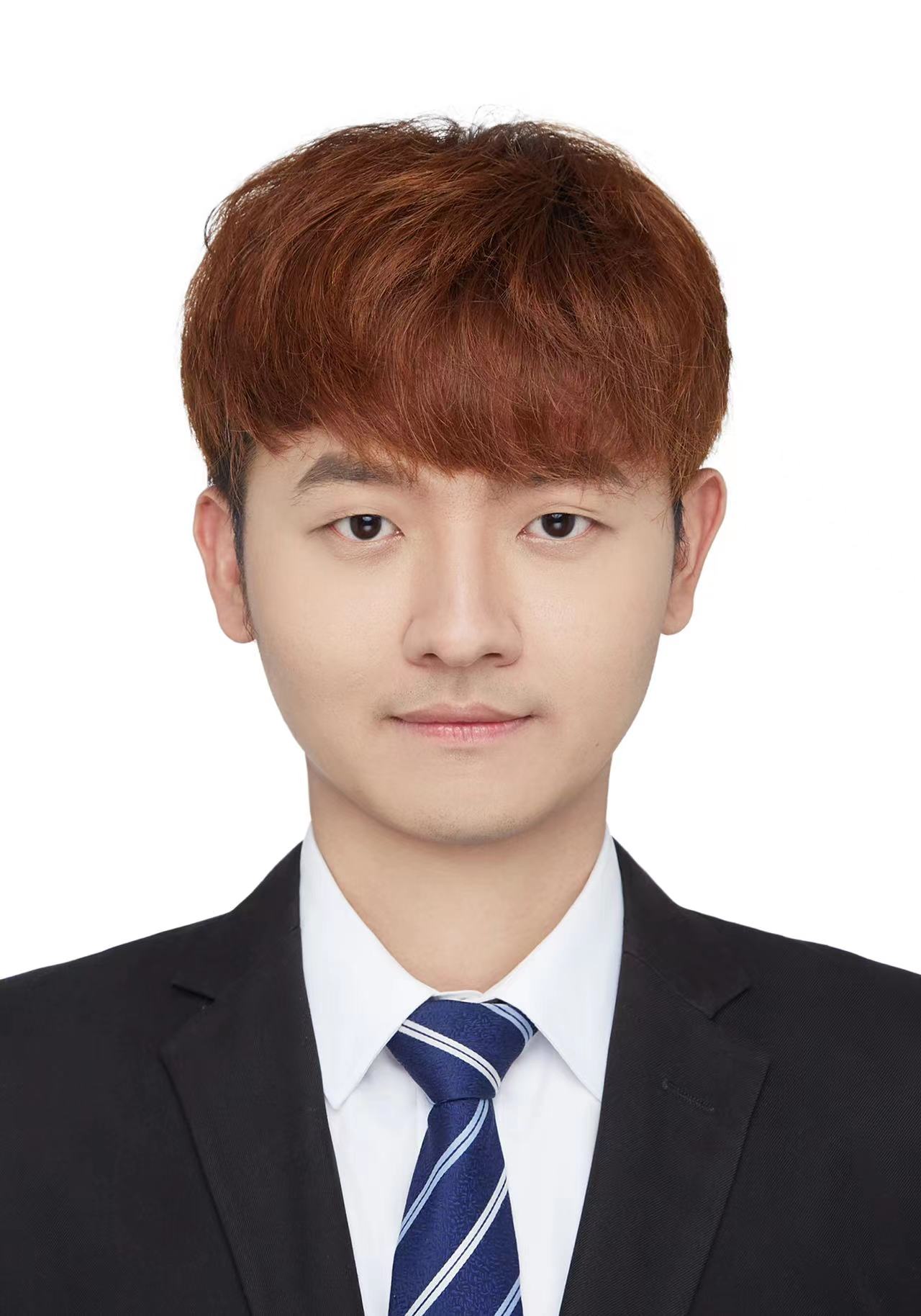}}]{Yongqi Tian}
received his B.S. degree from Beijing Institute of Technology, China, in 2019. He is currently pursuing a master’s degree in optical engineering at the Beijing Institute of Technology. His current research interests include Computer Vision and Deep Learning.
\end{IEEEbiography}

\begin{IEEEbiography}[{\includegraphics[width=1in,height=1.25in,clip,keepaspectratio]{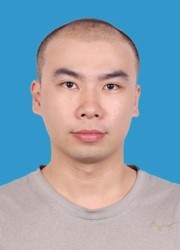}}]{Cong Lin}
is currently an assistant professor at the School of Intelligent Systems Science and Engineering, Jinan University, Zhuhai, China. He obtained his Ph.D. degree in Software Engineering from University of Macau in 2017. His research interests include pattern recognition, machine learning, and computer vision. He is currently working on applicational image to image transforms using deep learning and vision-based object detectors for practical use.
\end{IEEEbiography}

\begin{IEEEbiography}[{\includegraphics[width=1in,height=1.25in,clip,keepaspectratio]{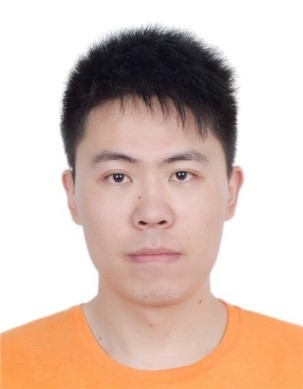}}]{Xinyuan Zhang}
received the B.S. and Ph.D. degrees in computer science from Sun Yat-sen University, Guangzhou, China, in 2014 and 2019, respectively. From 2015 to 2017, he was a Research Assistant with the Department of Electronic Engineering, City University of Hong Kong, Hong Kong, China. His current research interests include evolutionary computation algorithms, swarm intelligence algorithms, large-scale optimization their applications in real-world problems.
\end{IEEEbiography}

\begin{IEEEbiography}[{\includegraphics[width=1in,height=1.25in,clip,keepaspectratio]{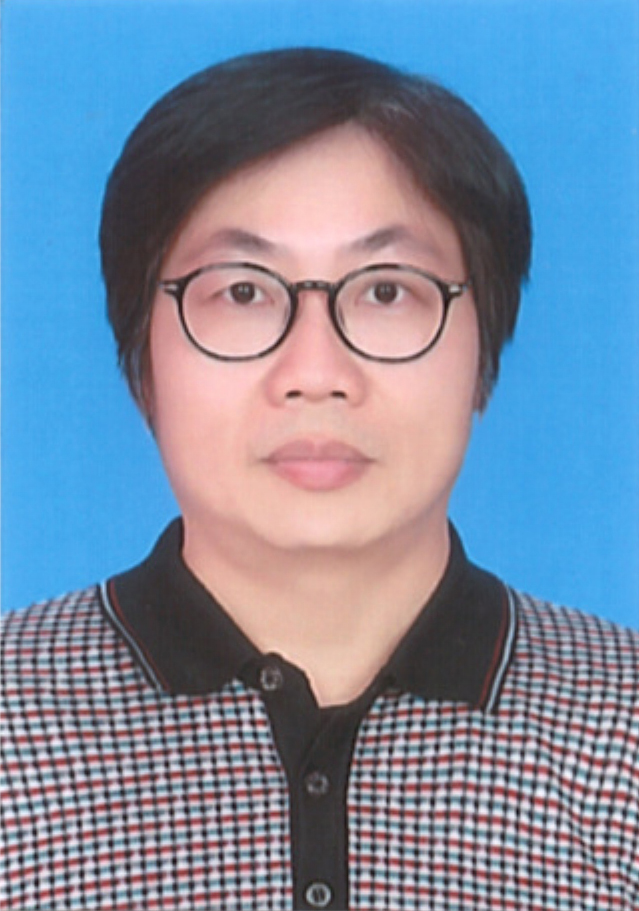}}]{Xiaoxiang Liu}
is currently an associate professor at the school of Intelligent Systems Science and Engineering, Jinan University, Zhuhai, China. He obtained his Ph.D. degree, M.Sc. degree and B.Sc. degree all in Aeronautical and Astronautical Manufacturing Engineering from Northwestern Polytechnical University. His research interests include Computer Vision and Artificial Intelligence.
\end{IEEEbiography}


%\vfill


\end{document}